\documentclass[10pt,twocolumn,letterpaper]{article}

\usepackage[pagenumbers]{iccv} 

\newcommand{\ms}{\scriptscriptstyle} 

\usepackage[svgnames]{xcolor}
\usepackage{booktabs}
\usepackage{makecell}
\usepackage{colortbl}%
  \newcommand{\myrowcolour}{\rowcolor[gray]{0.925}}
  \newcommand{\myrowcoloursub}{\rowcolor[gray]{0.8}}
  \newcommand{\highest}[1]{\textcolor{blue}{#1}}%
  \newcommand{\secondhighest}[1]{\textcolor{magenta}{#1}}

\definecolor{iccvblue}{rgb}{0.21,0.49,0.74}
\usepackage[pagebackref,breaklinks,colorlinks,allcolors=iccvblue]{hyperref}
\usepackage{caption}
\usepackage{subcaption}

\def\paperID{10864} 
\def\confName{ICCV}
\def\confYear{2025}

\title{From Prompt to Progression: Taming Video Diffusion Models for Seamless Attribute Transition}

\author{Ling Lo\\
National Yang Ming Chiao Tung University\\
{\tt\small linglo.ee08@nycu.edu.tw}
\and
Kelvin C.K. Chan\\
Google DeepMind\\
{\tt\small  kelvinckchan@google.com}
\and
Wen-Huang Cheng\\
National Taiwan University\\
{\tt\small wenhuang@csie.ntu.edu.tw}
\and
Ming-Hsuan Yang\\
UC Merced\\
{\tt\small mhyang@ucmerced.edu}
}

\begin{document}
\maketitle
\begin{abstract}
Existing models often struggle with complex temporal changes, particularly when generating videos with gradual attribute transitions.
The most common prompt interpolation approach for motion transitions often fails to handle gradual attribute transitions, where inconsistencies tend to become more pronounced.
%
%
In this work, we propose a simple yet effective method to extend existing models for smooth and consistent attribute transitions, through introducing frame-wise guidance during the denoising process.
Our approach constructs a data-specific transitional direction for each noisy latent, guiding the gradual shift from initial to final attributes frame by frame while preserving the motion dynamics of the video. 
Moreover, we present the Controlled-Attribute-Transition Benchmark (CAT-Bench), which integrates both attribute and motion dynamics, to comprehensively evaluate the performance of different models. 
We further propose two metrics to assess the accuracy and smoothness of attribute transitions. 
Experimental results demonstrate that our approach performs favorably against existing baselines, achieving visual fidelity, maintaining alignment with text prompts, and delivering seamless attribute transitions. Code and CATBench are released: \url{https://github.com/lynn-ling-lo/Prompt2Progression}

\end{abstract}    
\vspace{-4mm}
\section{Introduction}
\begin{figure}
\begin{center}
\includegraphics[width=\columnwidth]{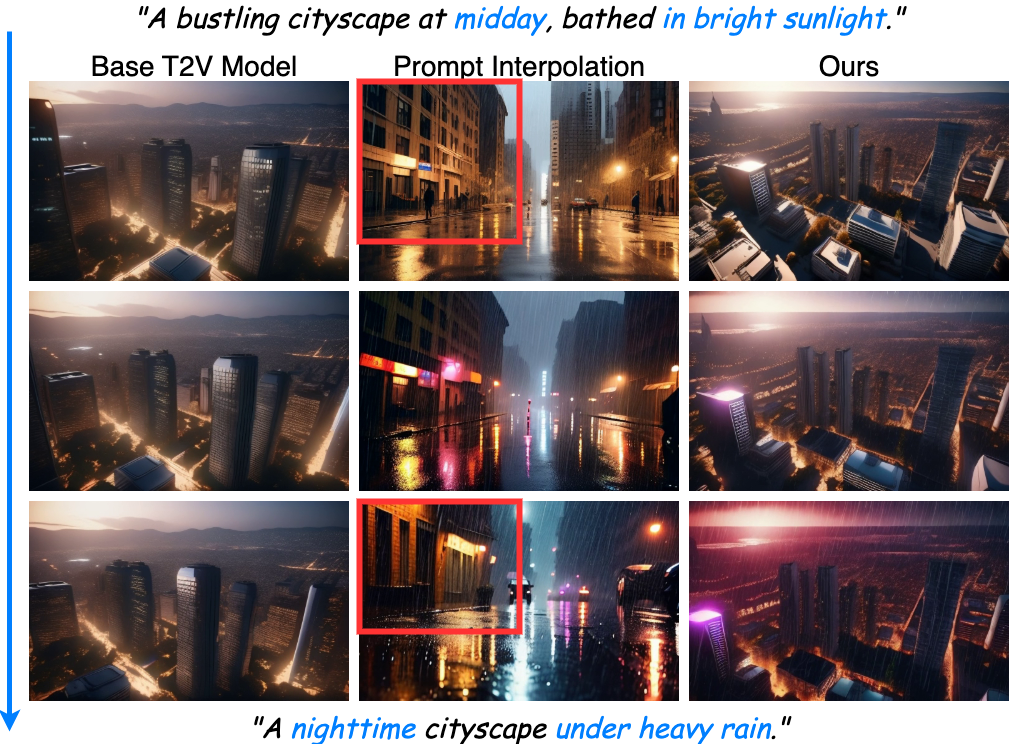}
\end{center}
\vspace{-4mm}
   \caption{Example of Video Generation with Attribute Transitions Using the Same Base Model. The base model generates static appearances throughout the video. Prompt interpolation leads to inconsistencies, such as abrupt changes in the buildings, while our method ensures smoother and more consistent attribute transitions.} 
   \vspace{-6mm}
\label{fig:intro}
\end{figure}

\vspace{-2mm}
Diffusion models have made remarkable advancements in text-to-video generation~\cite{lee2024grid, hong2022cogvideo, wu2023tune, he2022lvdm, chen2024videocrafter2, ma2024latte, guo2023animatediff, wang2023modelscope}. Compared to image generation, the additional time dimension in videos provides an opportunity to create more dynamic and engaging content. Recent developments in large-scale datasets and architecture designs have led to high-quality, powerful text-to-video (T2V) models that generate videos with remarkable coherence and realism.
Despite their success in producing temporally consistent video sequences, these models often struggle with complex variation over the time dimension, such as varying motions and attribute transitions. Real-world videos, on the other hand, are composed of sequences featuring different motions, color changes, varying light conditions, and evolving backgrounds over time. The limitation inevitably restricts the potential of video generation in practical applications, where the ability to handle dynamic changes is essential.

As shown in Figure~\ref{fig:intro}, when attempting to generate a transition from\textit{``A bustling cityscape at midday, bathed in bright sunlight.''} to \textit{``A nighttime cityscape under heavy rain.''} in a video, the base T2V model fails to produce meaningful attribute changes, resulting in a static appearance throughout the sequence. 
To address temporal variations, recent studies propose additional networks to facilitate multi-prompt video generation~\cite{ge2022long,villegas2022phenaki}, utilizing multiple descriptions that capture the desired transitions over time. However, they require costly retraining and rely on large, complex text-video datasets. 
To mitigate training cost, training-free methods~\cite{wang2023genl, qiu2023freenoise} are developed, generally employing prompt interpolation to improve transition fluidity. Nevertheless, they are primarily designed for dynamic motions, and the generation of smooth and gradual attribute transitions remains underexplored. Attribute transitions typically involve a more gradual process that unfolds over several frames. Consequently, inconsistencies from these methods become more pronounced, making them less effective for scenarios that demand smooth and seamless attribute changes. As observed in Figure~\ref{fig:intro}, prompt interpolation struggles to maintain object consistency, causing elements like buildings to distort across frames.

In this work, we focus on extending the base model to generate progressive attribute transitions without additional training. Our approach begins by crafting a \textbf{\textit{transitional direction}} representing the transition from the initial attribute to the final attribute\footnote{For example, in Figure~\ref{fig:intro}, the initial attribute is \textit{``midday, in bright sunlight''}, and the final attribute is \textit{``nighttime, under heavy rain''.}}.
Using the crafted transitional direction, we provide frame-wise guidance to the video throughout the denoising process to obtain smooth transitions of the desired attribute states. 
To stabilize generation, we propose \textbf{\textit{neutral prompt anchoring}}, which anchors the generation to a neutral prompt while applying the transitional direction. The neutral prompt mitigates inconsistencies in prompt interpolation, where discrepancies between the first and last prompts can lead to abrupt changes, ensuring a seamless transition throughout the video. Our method is training-free and achieves smooth, controlled attribute transitions without sacrificing motion dynamics and video quality.

To evaluate attribute transitions in video generation, we introduce a new benchmark, \textbf{\textit{Controlled-Attribute-Transition Benchmark (CAT-Bench)}}, designed to measure the smoothness and accuracy of attribute changes throughout the video. Specifically, we design prompts across eight categories, encompassing human, object, and background attribute transitions. Each prompt contains an attribute transition alongside subject motion, creating a complex scene that captures both attribute and motion dynamics. In addition, we develop two metrics: 1) \textit{Wholistic Transition Score} and 2) \textit{Frame-wise Transition Score} to assess the effectiveness and smoothness of attribute transitions in videos. We extensively benchmark multiple baselines, including single-prompt T2V generation and multi-prompt T2V generation. Our experiments show the effectiveness of our method in generating videos that smoothly align with the provided text prompts. 
The contributions of this work are:
\begin{itemize}
    \item We propose a simple yet effective approach for generating videos with smooth and gradual attribute transitions. Our training-free method integrates attribute changes without compromising the motion dynamics of the video.
    \item We introduce a new benchmark designed to evaluate the attribute transition in video generation. In addition, we propose two novel metrics to quantitatively evaluate the smoothness and fidelity of attribute transitions.
    \item Our experimental results demonstrate the effectiveness of the proposed approach. Moreover, we show that the proposed metrics are well aligned with human perception.
\end{itemize}

\section{Related Work}
\vspace{-2mm}
\paragraph{Single-Prompt Video Generation via Diffusion Models.}
Diffusion models have recently become a leading approach for T2V generation~\cite{wang2023modelscope,he2022lvdm,chen2024videocrafter2, guo2023animatediff, ma2024latte}.
Building on advances in large-scale text-to-image (T2I) models, ModelScope~\cite{wang2023modelscope} and VideoCrafter2~\cite{chen2024videocrafter2} incorporate additional spatial-temporal blocks for temporal dependencies.
AnimateDiff~\cite{guo2023animatediff} finetunes a pre-trained T2I model to adapt to motion patterns seamlessly. Latte~\cite{ma2024latte} uses a series of transformer blocks to effectively model video distributions. Although these models can generate high-quality videos from a single prompt, they often struggle with capturing dynamic changes across the temporal dimension.

\vspace{-4mm}
\paragraph{Multi-Prompt Video Generation via Diffusion Models.}
Since a single prompt is often insufficient to fully describe a video, Villegas~\etal introduce Phenaki~\cite{villegas2022phenaki} that takes multiple prompts as input. Although effective, retraining the model requires substantial computational resources and large amounts of data. In contrast, GenL~\cite{wang2023genl} and FreeNoise~\cite{qiu2023freenoise} propose training-free methods that leverage existing models to generate short clips using different prompts, combining them into a video with prompt interpolation for smooth transitions. However, they primarily address dynamic motions and struggle with attribute transitions, which demand a more gradual process that prompt interpolation fails to achieve. Additionally, VideoTetris~\cite{tian2024videotetris}, FreeBloom~\cite{huang2024freebloom}, and MEVG~\cite{oh2025mevg} use large language models (LLMs) to describe the video frame by frame. Yet for attribute transition, the subtle changes between frames are difficult to capture accurately with textual descriptions.

\vspace{-4mm}
\paragraph{Score Estimates Guidance.} 
As diffusion models can be seen as score-based generative models~\cite{song2021score}, research~\cite{ho2021classifierfree,dhariwal2021diffusion} that manipulates the score estimates has proven effective in guiding the generation process post-training, enabling finer control over the resulting outputs. Liu~\etal~\cite{liu2022compositional} combine multiple estimates to facilitate changes in image composition. Brack~\etal~\cite{brack2023sega} introduce semantic score guidance that scales monotonically, enabling fine-grain control over subtle semantic changes. However, these efforts have mainly focused on fine-grained control in T2I generation. Unlike static images, attribute transition in video demands continuous attribute changes across frames, presenting unique and complex challenges and remaining under-explored.

\section{Method}
\vspace{-2mm}
\begin{figure}
\begin{center}
\includegraphics[scale=0.2]{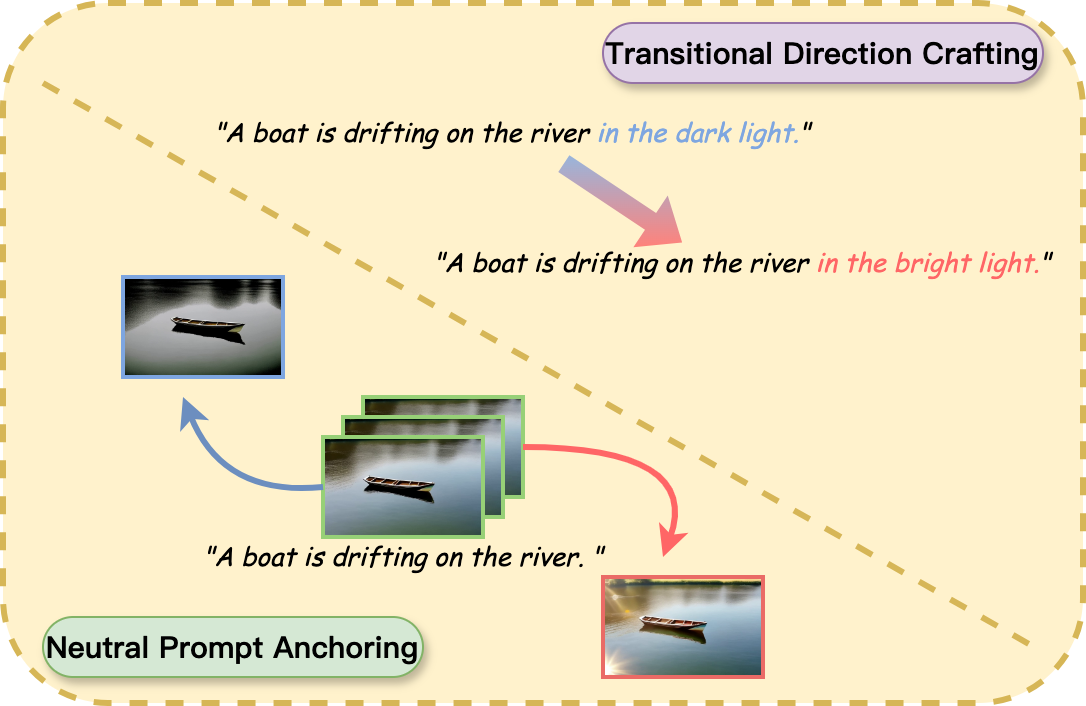}
\end{center}
\vspace{-4mm}
   \caption{Overview of Our Method. We first create the transitional direction, then guide the video sampled with a neutral prompt frame by frame to achieve smooth attribute transitions while maintaining motion dynamics.} 
\label{fig:method}
\vspace{-4mm}
\end{figure}
Given two descriptive prompts, $P_I$ and $P_F$, where $P_I$ describes the scene with initial attribute state\footnote{For example, \textit{``A boat is drifting on the river \textbf{in the dark light.}''}}, and $P_F$ specifies the scene with final attribute state\footnote{For example, \textit{``A boat is drifting on the river in \textbf{the bright light.}''}}, our goal is to generate a video that captures this attribute transition while maintaining consistent motion dynamics. 
We leverage a pre-trained T2V diffusion model and calibrate the sampling process during inference in a training-free manner. Figure~\ref{fig:method} illustrates an overview of the proposed method. 

The core idea is to first identify a transitional direction in the latent space, and then adjust the sampling process for the video latent, guiding each frame to follow the transition direction in a frame-wise manner. In this section, we start with a brief overview of T2V diffusion models, followed by a detailed description of our proposed method.

\subsection{Preliminaries}
\paragraph{T2V Diffusion Models.}
Text-to-video diffusion models aim to map a random noise vector $Z_T$ and a text condition $P$ to an output latent $Z_0$ through a denoising network $\epsilon_\theta$. 
During training, a forward diffusion process is first applied to transform the initial latent vector $Z_0\in R^{F\times C\times H\times W}$ into a noisy approximation $Z_t$ over timestep $t$, where $F$ represents the number of frames in the video, and $H$, $W$ and $C$ denote the height, width, and channel, respectively. Conditioning on $P$, the denoising network $\epsilon_\theta$ is trained to estimate the noise $\epsilon_t$ in each time step $t$ within the noisy video latent $Z_t$ by minimizing the objective:
\begin{equation}
    \min_{\theta}E_{Z_0, \epsilon\sim N(0,1), t\sim uniform(1,T)}\lVert \epsilon_t - \epsilon_\theta (Z_t,P)\rVert^2_2 .
\end{equation}
Since the objective is noise matching over multiple levels, the learned denoising score $\epsilon_\theta (Z_t, P)$ estimates the gradient of the log-density of the distribution of the noisy data $Z_t$ conditioning on $P$~\cite{song2021score}: 
\begin{equation}
    \epsilon_\theta (Z_t, P) \approx - \sigma_t \nabla_{Z_t} \log P(Z_t|P),
    \label{equ:log}
\end{equation}
where $P(Z_t|P)$ represent the distribution generated by the diffusion model conditioned on $P$.

\paragraph{Sampling from Diffusion Model.}
Classifier-free guidance allows diffusion models to trade off diversity for fidelity~\cite{ho2021classifierfree}. When a conditional and an unconditional diffusion model are jointly trained, the predicted noise during inference can be guided by combining both the conditional and unconditional denoising scores: 
\begin{equation}
    \tilde{\epsilon}_\theta (Z_t, P) = (1+\omega)\cdot\epsilon_\theta (Z_t, P) - \omega\cdot\epsilon_\theta (Z_t),
    \label{eqa: cfg}
\end{equation}
where $\omega$ is the guidance scale that controls the trade-off between mode coverage as well as sample fidelity.


\subsection{Transitional Direction Crafting}
\label{sec: transitional direction crafting}
We aim to generate a video, where each frame shifts progressively from displaying $P_I$ to $P_F$. First, we identify a transitional direction in the latent space that transitions from the $P_I$ to $P_F$, which involves increasing the likelihood of $P_F$ and decreasing the likelihood of $P_I$ in $Z_t$:
\begin{equation}
    \frac{P_\theta(P_F|Z_t)}{P_\theta(P_I|Z_t)}.
\end{equation}
By the Bayes' Rule, $P_\theta(P_F|Z_t)=\frac{P_\theta(Z_t|P_F)\cdot P_\theta(P_F)}{P(Z_t)}$, and assume a uniform prior over $P$, which causes the $P_\theta(P_F)$ and $P_\theta(P_I)$ terms to cancel, we have 
\begin{equation}
\frac{P_\theta(P_F|Z_t)}{P_\theta(P_I|Z_t)} \propto \frac{P_\theta(Z_t|P_F)}{P_\theta(Z_t|P_I) }.
\end{equation}
%
The direction for the transition can be interpreted as the gradient of the log-likelihood:
\begin{equation}
    \nabla_{Z_t} \log P(Z_t|P_F) -\nabla_{Z_t} \log P(Z_t|P_I).
    \label{eqa: gradient}
\end{equation}
Combining Equation~\ref{equ:log} and Equation~\ref{eqa: gradient}, we obtain the unit transitional direction in the $\epsilon$-space:
\begin{equation}
   D_{\epsilon_\theta}(Z_t, P_{I-F}) = \frac{\epsilon_\theta (Z_t, P_F) - \epsilon_\theta (Z_t, P_I)}{\lVert \epsilon_\theta (Z_t, P_F) - \epsilon_\theta (Z_t, P_I) \rVert_2}.
\end{equation}
The obtained $D_{\epsilon_\theta}(Z_t, P_{I-F})$ is a local, data-dependent direction that guides the transition from $P_I$ to $P_F$ for $Z_t$.

%
The quality of the transitional direction depends on the precision of the denoising score. Using simple terms in $P_I$ and $P_F$ may introduce ambiguity, leading to unintended changes as shown in Figure~\ref{fig:ablation_full}. 
To reduce such issues, we use full descriptive prompts instead of isolated attributes, ensuring the denoising score captures the intended context and enhances the accuracy of the transitional direction.

\subsection{Neutral Prompt Anchoring}
To achieve smooth attribute transitions in video generation, we manipulate the video latents during the denoising process at inference time. By guiding each frame along the transitional direction, we ensure a gradual and seamless change of attributes across the entire sequence. 

We leverage the transitional direction ${D}$ in the $\epsilon$-space and adjust the denoising score in Equation~\ref{eqa: cfg} for each frame. Notably, the transitional direction is crafted with a holistic view of the video, providing a consistent manipulation scale across all frames. Moreover, the direction of each frame can be treated independently, allowing fine-grained control over attribute adjustments throughout the video sequence. 
A straightforward approach is to incrementally add frame-dependent scales of $D$ to the score associated with $P_I$, $\epsilon_\theta (Z_t, P_I)$. The refined denoising score of the $j$th frame of the video can be formulated as:
\begin{equation}
    \hat{\epsilon}_\theta^j(Z_t, P_{I-F}) = \epsilon_\theta^j(Z_t, P_I)+\alpha_j \cdot D_{\epsilon_\theta}^j(Z_t, P_{I-F}),
    \label{eqa: sample}
\end{equation}
where $\hat{\epsilon}_\theta^j(Z_t, P_{I-F})$ represents the denoising score of the $j$th frame with specified attribute transition and the scale factor $\alpha_j$ changes uniformly over the sequence of frames.

However, as shown in Figure~\ref{fig:ablation}, inconsistencies are observed in the synthesized video, where the final frame differs significantly from the first due to the distance effect~\cite{shen2020interpreting}: the transitional direction defines a hyperplane separating the initial and final attributes. Denoising scores near this boundary effectively yield smooth and controlled transitions. However, moving too far from the hyperplane can amplify changes, causing unwanted distortions.

As such, we employ a neutral prompt to keep the denoising score near the boundary throughout the denoising process. Specifically, given $P_I$ and $P_F$, we construct a neutral prompt $P_N$\footnote{For example, \textit{``A boat is drifting on the river.''} in Figure~\ref{fig:method}.} by extracting the common elements from both prompts that describe only the motion in the video without referencing the attributes we intend to change. We designate the middle frame to have a neutral attribute. The transition can be achieved by adjusting the denoising score along or against the transitional direction for the following and preceding frames. Equation~\ref{eqa: sample} can thus be rewritten as:
\begin{equation}
    \small
    \hat{\epsilon}_\theta^j(Z_t,P_{\ms I \text{-} F}) =
    \begin{cases} 
        \epsilon_\theta^j(Z_t, P_N) - \alpha_{\ms M \text{-} j}\cdot D_{\ms \epsilon_\theta}^j(Z_t, P_{\ms I \text{-} F}), & 
        j < M \\ 
        \epsilon_\theta^j(Z_t, P_N), & 
        j = M\\ 
        \epsilon_\theta^j(Z_t, P_N) + \alpha_{\ms j \text{-} M} \cdot D_{\ms \epsilon_\theta}^j(Z_t, P_{\ms I \text{-} F}), & 
        j > M 
    \end{cases},
\end{equation}
where $M$ is the middle frame of the video. 
By anchoring the middle frame with a neutral prompt, which is devoid of dominant attributes and thus near the boundary, we keep the transition near the hyperplane, effectively mitigating the distance effect.

\subsection{Overall Sampling Flow}
As early stages of the denoising process mainly establish geometric, such as the layout of the video~\cite{balaji2022ediff,hertz2022prompt,cao2023masactrl,qiu2023freenoise}, in the first $\tau$ denoising steps, we employ the neutral prompt and sample from diffusion model using classifier-free guidance in Equation~\ref{eqa: cfg} to construct a general video structure for enhanced consistency. 
After completing $\tau$ denoising steps, we guide the sampling process using the transition direction to achieve smooth attribute transitions. It is worth noting that classifier-free guidance can still be applied to enhance the quality of the generated samples:
\begin{equation}
    \tilde{\epsilon}_\theta(Z_t, P_{I-F}) = (1+\omega)\cdot \hat{\epsilon}_\theta (Z_t, P_{I-F}) - \omega\cdot\epsilon_\theta(Z_t).
\end{equation}

\section{CAT-Bench}
\vspace{-2mm}
To assess model performances in generating videos with attribute transitions, we construct a new benchmark, Controlled-Attribute-Transition Benchmark (CAT-Bench), to evaluate the fidelity of models in handling attribute transitions while maintaining motion consistency and fluidity.

\subsection{Benchmark Construction}
We define \textbf{eight} attribute categories for transitions: four focused on \textit{human attributes} (age, beard, makeup, and hair), two on \textit{subject attributes} (color and material), and two on \textit{background attributes} (light conditions and weather conditions).
We generate prompt pairs to capture attribute changes in videos. The first prompt describes the initial frame, while the second outlines the final frame. Each pair is carefully designed to incorporate both motion and attribute transitions. To maintain consistent motion throughout the video, both prompts are identical except for the part specifying the evolving attribute, allowing each prompt pair to convey ongoing motion and attribute changes seamlessly. 

We use GPT-4~\cite{achiam2023gpt} to generate multiple samples across all attribute categories, producing 15 prompt pairs per category, resulting in a total of 120 prompts. To facilitate a comprehensive assessment, we also use GPT-4 to generate a single prompt describing the full attribute transition for each prompt pair, allowing us to compare and analyze different generation approaches thoroughly. 

\subsection{Evaluation Metrics}

We introduce two metrics to assess attribute transition quality: \textit{Wholistic Transition Score} and \textit{Frame-wise Transition Score}. \textit{Wholistic Transition Score} evaluates the overall correctness of the attribute transition, determining whether the change from the first to the last frame follows the intended progression. The score is computed using a CLIP-based directional similarity between the transformation of visual features from the initial to the final frame and the semantic change described by the prompts:
\begin{equation}
    \frac{\langle E_I(F^l) - E_I(F^f), E_T(P^s) - E_T(P^f)\rangle}{\lVert E_I(F^l) - E_I(F^f)\rVert \lVert  E_T(P^s) - E_T(P^f) \rVert},
\end{equation}
where $E_I$ denotes the CLIP image encoder and 
$E_T$ represents the CLIP text encoder. 
$F^f$ and 
$F^l$ are the visual features of the first and last frames, respectively, while $P^f$ and $P^l$ are the features of the first and second prompts. 

In contrast, \textit{Frame-wise Attribute Transition} assesses the smoothness and gradual nature of the transition, ensuring that each frame changes consistently without abrupt shifts. We compute the directional similarity between every consecutive frame in the video to determine if each pair aligns with the direction specified in the prompt pair:
\begin{equation}
    \frac{1}{N-1} \sum_{i=1}^{N-1} \frac{\langle E_I(F^{i+1}) - E_I(F^i), E_T(P^s) - E_T(P^f) \rangle}{\lVert E_I(F^{i+1}) - E_I(F^i) \rVert \lVert E_T(P^s) - E_T(P^f) \rVert} ,
\end{equation}
thereby verifying the smoothness of the transition.

\section{Experimental Results} 
\vspace{-2mm}
We conduct experiments to evaluate the performance of our method quantitatively and qualitatively. 
\subsection{Experimental Setup} 
Our method is built on the open-source T2V foundation model, VideoCrafter2~\cite{chen2024videocrafter2}, which serves as the backbone of our implementation.
In our experiments, we set the total number of denoising time steps $T$ to 50, consistent with the backbone model. The number of frames $F$ in each generated video is set to 32. The parameter $\tau$, which controls the sampling process, is set to 5.

\paragraph{Baselines.}
We evaluate our method alongside various state-of-the-art baselines, including single-prompt and multiprompt approaches. The single-prompt baselines consist of AnimateDiff~\cite{guo2023animatediff}, ModelScope~\cite{wang2023modelscope}, Latte~\cite{ma2024latte}, and VideoCrafter2~\cite{chen2024videocrafter2}, which are foundational T2V models. The multi-prompt baselines include FreeBloom~\cite{huang2024freebloom}, Gen-L~\cite{wang2023genl}, Freenoise~\cite{qiu2023freenoise}, and VideoTetris~\cite{tian2024videotetris}. Similar to our method, Gen-L and Freenoise are training-free approaches and also use VideoCrafter2 as their backbone. 

\paragraph{Benchmarks.} In addition to our CAT-Bench, we evaluate the effectiveness of attribute transition using TC-Bench-T2V~\cite{feng2024tc}. TC-Bench is a benchmark designed to assess temporal compositionality in video generation, featuring different types of object state changes. For our evaluation, we focus on the attribute and background shifting subsets to assess whether the transition can be fulfilled effectively. While TC-Bench provides insight on various object state transitions, it lacks scenes with motion dynamics in prompts involving attribute and background transition. The limitation underscores the importance of our CAT-Bench, which addresses the gap by incorporating motions to evaluate attribute transitions in complex scenes.

\subsection{Qualitative Evaluation}
\begin{figure*}
\begin{center}
\includegraphics[scale =0.085]{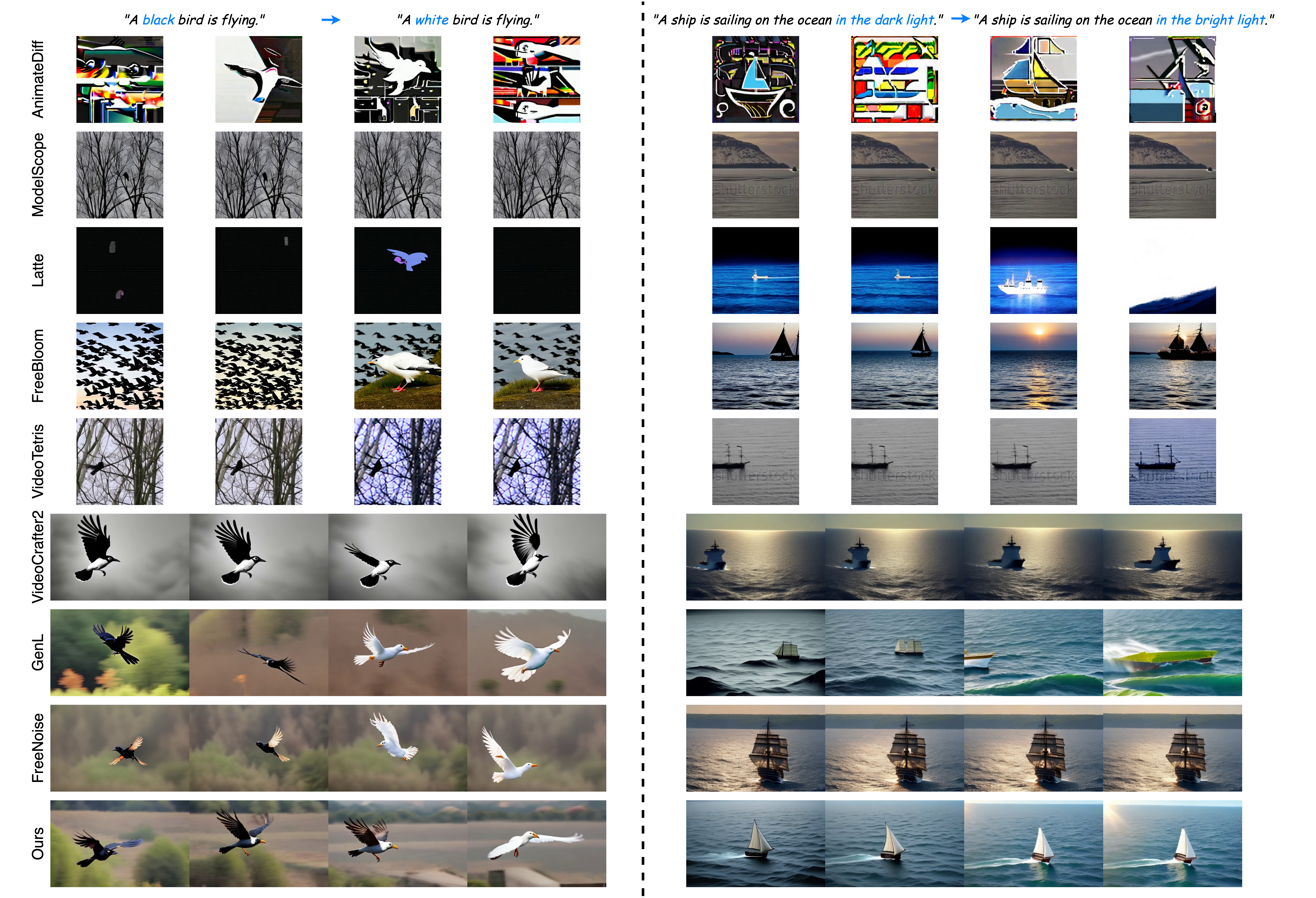}
\end{center}
\vspace{-4mm}
   \caption{Generation Results for the Specified Attribute Transition. AnimateDiff, ModelScope, Latte, and VideoCrafter2 are foundational text-to-video (T2V) models. FreeBloom and VideoTetris use LLMs to generate multiple detailed prompts as input and generate videos. Gen-L, FreeNoise, and our method are training-free approaches and utilize VideoCrafter2 as the backbone.} 
\label{fig: qualitative}
\end{figure*}

Figure~\ref{fig: qualitative} showcases the generation results from baseline approaches and the proposed method. Results from AnimateDiff, ModelScope, Latte, and VideoCrafter2 show they struggle to handle attribute transitions effectively when guided by a single prompt. Without retraining on relevant data, it is challenging to interpret or execute the attribute transitions specified in the prompts, resulting in subpar performance. On the other hand, FreeBloom and VideoTetris utilize LLMs to generate one descriptive prompt per frame. However, because subtle attribute state differences between consecutive frames are difficult to convey accurately through text, their results exhibit abrupt attribute shifts or unwanted changes. GenL and FreeNoise use prompt interpolation to achieve transitions between motions, but their approach often fails to produce intermediate attribute states reliably, indicating that interpolation does not always translate well into intermediate attributes. Additionally, because the first and last frames are generated from different prompts, the frames are likely to appear as disjointed images, introducing further inconsistency. In contrast, our proposed framework generates videos with smooth and accurate attribute transitions, ensuring consistency and a gradual progression across frames. Moreover, our method achieves the desired attribute transitions while maintaining video quality comparable to the base model. Quantitative evaluations of video quality are reported in the supplementary materials, along with additional qualitative results.

\subsection{Quantitative Evaluation}
\begin{table*}[ht]
\centering
\small
\caption{Evaluation Results of the Wholistic Transition Score on CAT-Bench. The highest scores are highlighted in \highest{blue}.}
\label{tab: wholistic}
\vspace{-2mm}
\begin{tabular}{lccccccccc}
\toprule
 & Age & Beard & Makeup & Hair & Color & Material & Light & Weather & Overall\\
\myrowcoloursub
\multicolumn{10}{l}{\textit{Single-Prompt Generation}}
\\

AnimateDiff& 0.0143 & 0.0012 & 0.0133 & 0.0098 & 0.0001 & -0.0156 & -0.0021 & 0.0207&0.0082\\
\myrowcolour
ModelScope& 0. 0201 & -0.0170 & -0.0025 & -0.0137 & 0.0285 & 0.0009 & 0.0186 & -0.0047 &0.0042\\

Latte& 0.0079 & 0.0015 & -0.0006 & -0.0218& 0.0210 & 0.0012 & 0.0003 & 0.0046 &0.0019\\
\myrowcolour
VideoCrafter2& 0.0002 & 0.0019 & 0.0178 & 0.0061 & -0.0063 & -0.0169 & -0.0084 & 0.0041 &0.0022\\
\myrowcoloursub
\multicolumn{10}{l}{\textit{Multi-Prompt Generation}}
\\
Free-Bloom& 0.1561 & 0.0255 & 0.0718 & 0.2244 & 0.0031 & 0.0162 & 0.0746 & 0.1987& 0.1077\\
\myrowcolour
VideoTetris& 0.0061 & 0.0210 & -0.0074 & 0.0295 & 0.0222& -0.0008 & 0.0272& -0.0052 & 0.0134\\

Gen-L& 0.1902 & 0.0015 & 0.0891 & 0.1881 & 0.0439 & -0.0033 & 0.0769 & 0.2264 &0.1166\\
\myrowcolour
FreeNoise& 0.1085 & 0.0003 & 0.0586 & 0.1142 & 0.0104 & -0.0037 & 0.0297 & 0.0834&0.0578\\

\textbf{Ours}& \highest{\textbf{0.2257}} &\highest{\textbf{0.0332}} & \highest{\textbf{0.1228}} & \highest{\textbf{0.2440}}& \highest{\textbf{0.0538}} & \highest{\textbf{0.0284}} & \highest{\textbf{0.1127}} & \highest{\textbf{0.2480}} &\highest{\textbf{0.1486}}\\
\bottomrule
\end{tabular}
\end{table*}

\begin{table*}[ht]
\centering
\small
\caption{Evaluation Results of the Frame-wise Transition Score on CAT-Bench. The highest scores are highlighted in \highest{blue}.}
\vspace{-2mm}
\label{tab:framewise}
\begin{tabular}{lccccccccc}
\toprule
 & Age & Beard & Makeup & Hair & Color & Material & Light & Weather & Overall\\
\myrowcoloursub
\multicolumn{10}{l}{\textit{Single-Prompt Generation}}
\\
AnimateDiff& -0.0002 & 0.0007 & 0.0011 & 0.0001& 0.0002 & 0.0009& -0.0002 & 0.0021 &0.0004\\
\myrowcolour
ModelScope& 0.0010 & 0.0003 & 0.0001 & -0.0016& 0.0020 & 0.0001& 0.0028 & -0.0009 & 0.0001\\

Latte& 0.0009 & -0.0009 & -0.0001 & -0.0033 & 0.0007 & -0.0001& -0.0001& 0.0019 &-0.0002\\
\myrowcolour
VideoCrafter2& 0.0005 & 0.0011 & 0.0018 & 0.0001 & -0.0012 & -0.0014 & -0.0013& 0.0010 & 0.0003\\
\myrowcoloursub
\multicolumn{10}{l}{\textit{Multi-Prompt Generation}}
\\
Free-Bloom& 0.0119 & 0.0026 & 0.0047 & -0.0153 & 0.0002 & -0.0012 & 0.0071& -0.0169 &-0.0020\\
\myrowcolour
VideoTetris& 0.0013 & -0.0002 & 0.0012 & 0.0024 & 0.0022 & 0.0009 & 0.0011 & 0.0005 &0.0012\\
Gen-L& 0.0262 & -0.0010& 0.0120 & 0.0246 & 0.0056 & 0.0001 & 0.0095& 0.0271&0.0135\\
\myrowcolour
FreeNoise& 0.0156 & 0.0005 & 0.0066 & 0.0153 & 0.0009 & -0.0012 & 0.0033 & 0.0086 &0.0066\\

\textbf{Ours}& \highest{\textbf{0.0321} }& \highest{\textbf{0.0044}} & \highest{\textbf{0.0237}} & \highest{\textbf{0.0382}} & \highest{\textbf{0.0073}} & \highest{\textbf{0.0043}} & \highest{\textbf{0.0157}} & \highest{\textbf{0.0306}} &\highest{\textbf{0.0201}}\\
\bottomrule
\end{tabular}
\end{table*}

\paragraph{CAT-Bench.} Table~\ref{tab: wholistic} presents the Wholistic Transition Score evaluation results on CAT-Bench. The table shows that single-prompt methods achieve lower scores compared to multi-prompt-based ones, indicating that single-prompt approaches struggle to produce accurate attribute transitions. The observation aligns with the qualitative evaluation results. Although multi-prompt approaches generally perform better, prompt interpolation approaches, Gen-L and FreeNoise, show suboptimal results for transitions involving \textit{beard}, which are described as \textit{with beard} and \textit{without beard} in prompts. The results indicate that interpolating between binary attributes specified using ``with'' and``without'' fails to generate intermediate attribute states effectively. 

Table~\ref{tab:framewise} reports the Frame-wise Transition Score, which measures the smoothness of attribute transitions. It reveals that while some multi-prompt methods achieve correct wholistic transitions, they lack smoothness, as indicated by several negative values. The results suggest that the frame-wise transitions occasionally move against the desired direction, disrupting the smoothness of the video. Regarding correctness and smoothness, Tables~\ref{tab: wholistic} and Table~\ref{tab:framewise} show that our method demonstrates improved performance on both criteria compared to existing approaches.

\begin{table}
\centering
\small
\caption{Quantitative Evaluation on TC-Bench-T2V. We report assertion-based TCR and TC-Score proposed in the benchmark. The highest score is highlighted in \highest{blue}.}
\vspace{-2mm}
\label{tab:TCbench}
\begin{tabular}{lcccc}
\toprule
& \multicolumn{2}{c}{Attribute} & \multicolumn{2}{c}{Background}\\
\myrowcoloursub
Method & TCR$\uparrow$ & TC-Score$\uparrow$ & TCR$\uparrow$ & TC-Score$\uparrow$\\
\midrule
ModelScope & 3.52 & 0.5942 & 3.54 & 0.5715\\
\myrowcolour
VideoCrafter2 & 4.25 & 0.6166 & 7.06 & 0.6338\\
FreeBloom & 6.32& 0.6256 & 24.02& 0.7394\\
\myrowcolour
TC-Bench & 13.08 & 0.6579&35.43&0.7916\\
\textbf{Ours} & \textbf{\highest{13.47}} & \textbf{\highest{0.6889}} & \textbf{\highest{35.88}} & \textbf{\highest{0.8043}}\\
\bottomrule
\end{tabular}
\end{table}

\begin{table}
\centering
\small
\caption{User Study. Participants are required to select the best video based on five different criteria. The table reports the percentage of users who choose each approach as the best.}
\vspace{-2mm}
\label{tab:human}
\begin{tabular}{lcccc}
\toprule
\myrowcoloursub
& VideoCrafter2 & GenL & Freenoise & \textbf{Ours}\\
\makecell{Transition\\Correctness}& 4.87\% & 28.91\% & 17.07\% & \highest{\textbf{49.12\%}}\\
\myrowcolour
\makecell{Transition\\Smoothness}& 7.69\% &  28.32\% & 18.53\% & \highest{\textbf{45.45\%}}\\
\makecell{Video\\Consistency} & 15.03\% & 27.62\% & 19.93\% & \highest{\textbf{37.41\%}}\\
\myrowcolour
\makecell{Motion\\Correctness}& 11.22\% & 29.08\% &  17.19\% & \highest{\textbf{41.75\%}}\\
\makecell{Overall\\Performance}& 5.81\%& 28.72\% & 19.27\% & \highest{\textbf{46.19\%}}\\
\bottomrule
\end{tabular}
\vspace{-4mm}
\end{table}

\paragraph{TC-Bench-T2V.}
Table~\ref{tab:TCbench} shows the evaluation results on TC-Bench-T2V. TCR and TC-Score are assertion-based evaluations conducted by LLMs, measuring transition completion and object consistency. TC-bench-T2V evaluates various object changes, including complex state changes such as a rose transitioning from deep crimson to fully bloomed. Our approach performs well on TC-Bench-T2V, demonstrating its effectiveness in handling diverse attribute transitions while maintaining object consistency.
\vspace{-4mm}
\paragraph{User Study.}

\begin{figure}
\begin{center}
\includegraphics[width=\columnwidth]{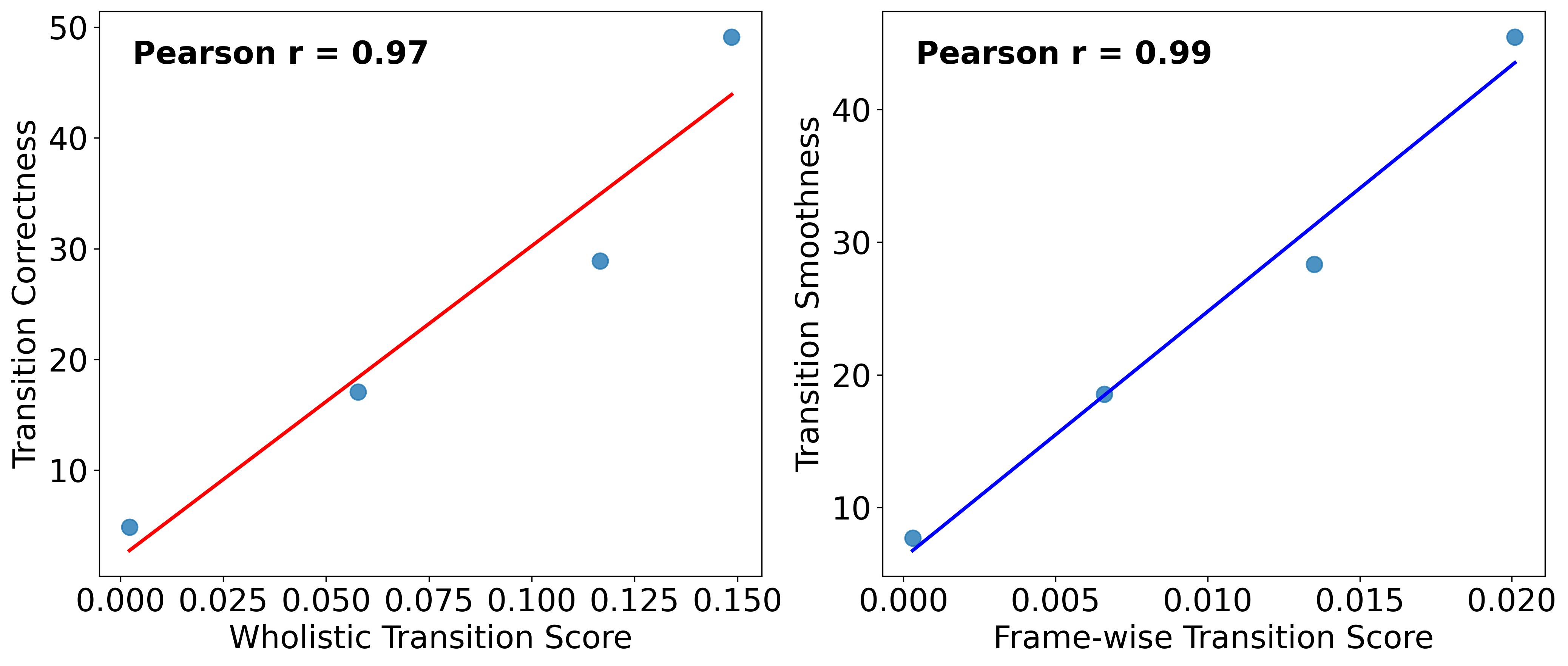}
\end{center}
\vspace{-4mm}
   \caption{Correlation between Our Metrics and Human Perception. The x-axis represents the Wholistic and Frame-wise Transition Score, while the y-axis corresponds to Transition Correctness and Smoothness from the user study. The results indicate a strong correlation, validating the effectiveness of our proposed metrics.} 
\label{fig:correlation}
\vspace{-4mm}
\end{figure}

We conduct a user study to evaluate the results based on human subjective assessments. Participants are asked to watch videos generated by four approaches that use VideoCrafter2 as the basemodel and select the best video based on five criteria: Transition Correctness, Transition Smoothness, Video Consistency, Motion Correctness, and Overall Performance. Table~\ref{tab:human} presents the results, showing that our method performs favorably compared to other approaches using the same base model. Moreover, the ranking orders for Transition Correctness and Smoothness align with the rankings in Tables~\ref{tab: wholistic} and Table~\ref{tab:framewise}, which report the Wholistic and Frame-wise Transition Scores. We illustrate the alignment in Figure~\ref{fig:correlation}, which demonstrates that our proposed metrics effectively correlate with human perception. The Pearson correlation coefficients (r) further confirm the strong correlation, validating the reliability of our metrics in assessing perceptual quality.

\begin{figure}
\begin{center}
\includegraphics[width=0.9\columnwidth]{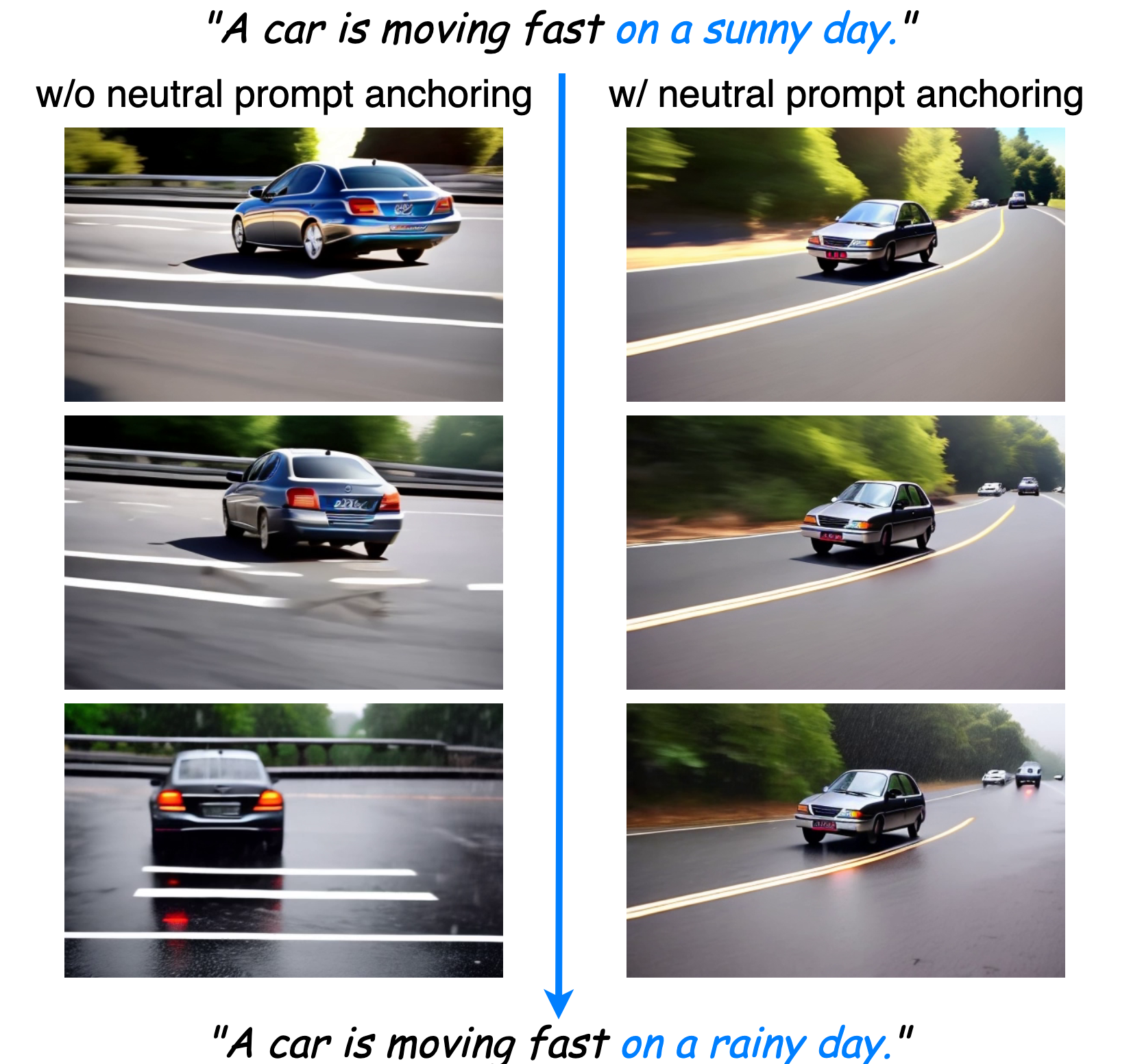}
\end{center}
\vspace{-6mm}
   \caption{Ablation on Neutral Prompt Anchoring. The unanchored approach introduces severe inconsistencies, such as unwanted changes in color and motion.} 
\label{fig:ablation}
\vspace{-4mm}
\end{figure}

\subsection{Ablation Study}
\textbf{Neutral Prompt.}
We perform an ablation study on neutral prompt anchoring to show its necessity. As depicted in Figure~\ref{fig:ablation}, without the neutral prompt, while the desired change is achieved, unintended alterations occur: the color of the car changes from blue to black and veers instead of continuing straight. In contrast, with neutral prompt anchoring, the appearance of the car is preserved, while the background transitions smoothly from sunny to rainy.

\paragraph{Full Prompt.}
\begin{figure}
\begin{center}
\includegraphics[width=\columnwidth]{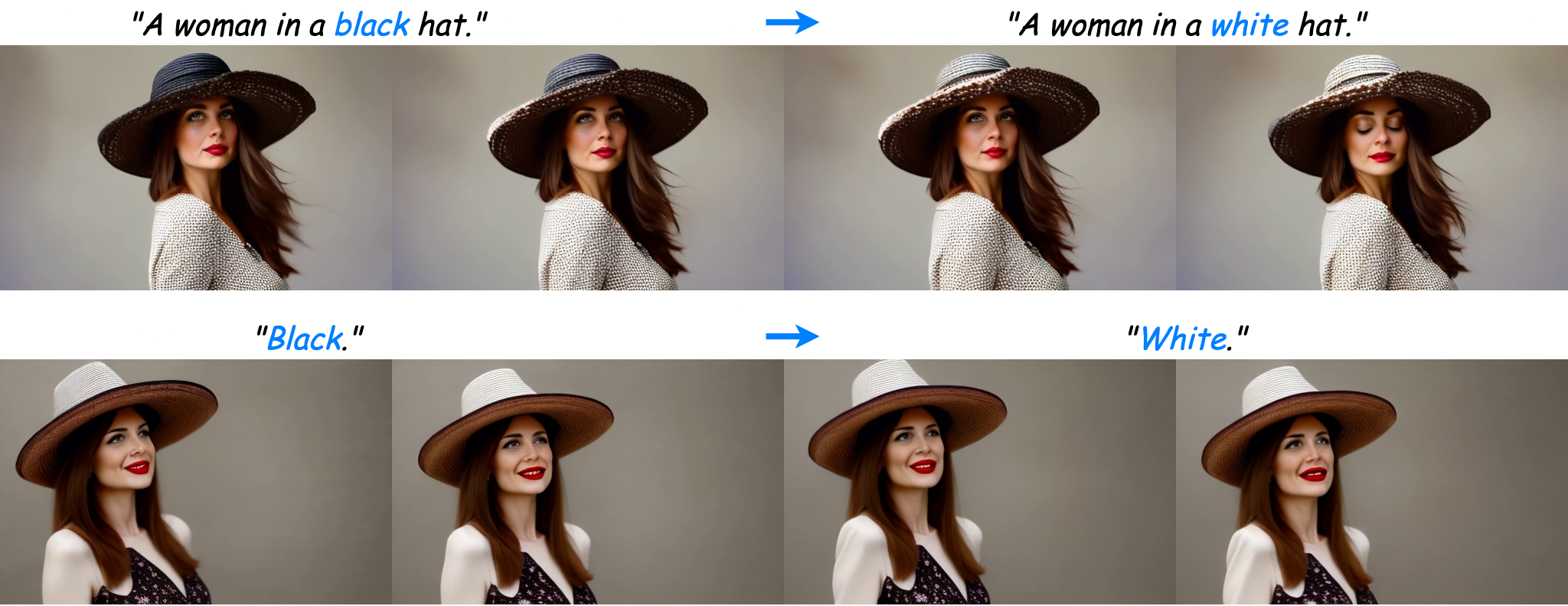}
\end{center}
\vspace{-4mm}
   \caption{Ablation on Full Prompt. Simple prompts lack context, leading to ambiguous or incomplete transitions. In contrast, full prompts provide clear guidance for precise attribute transitions.} 
\label{fig:ablation_full}
\vspace{-4mm}
\end{figure}

We conduct an ablation study on the use of full prompts. As discussed in Section~\ref{sec: transitional direction crafting}, the quality of transitional scenes depends on the conditional denoising score, which aligns more accurately with precise prompts. As shown in Figure~\ref{fig:ablation_full}, using a simple prompt such as ``\textit{black}'' and ``\textit{white}'' is ambiguous, as it may refer to different entities, leading to unintended changes or no change at all due to uncertainty. In contrast, using full prompts, such as ``\textit{a woman in a black hat.}'' transitioning to ``\textit{a woman in a white hat.}'', provides clear contextual guidance, resulting in a more precise and controlled transition.

\subsection{Broader Applicability}
\begin{figure}
\begin{center}
\subfloat[ Detailed Attribute Control.]{
\includegraphics[clip,width=\columnwidth]{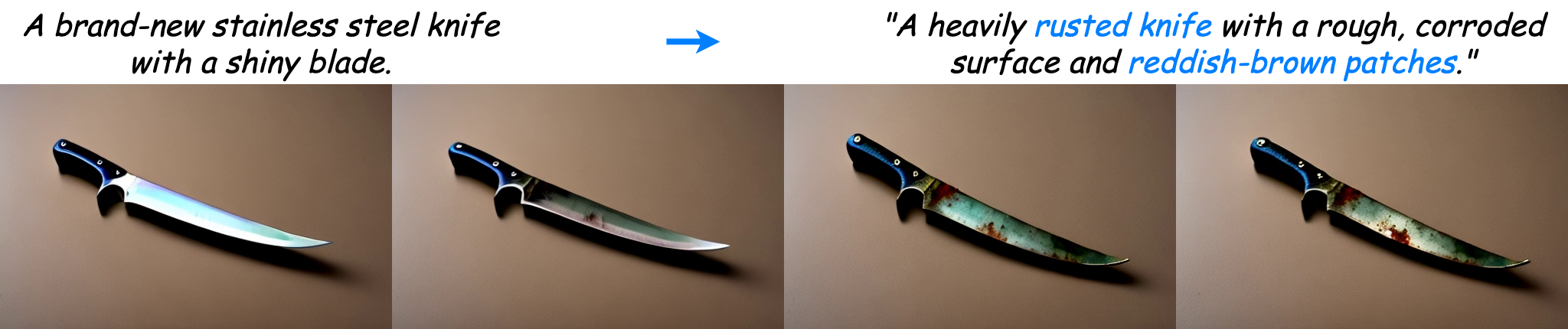}
}

\subfloat[Multiple Attributes Control.]{
\includegraphics[clip,width=\columnwidth]{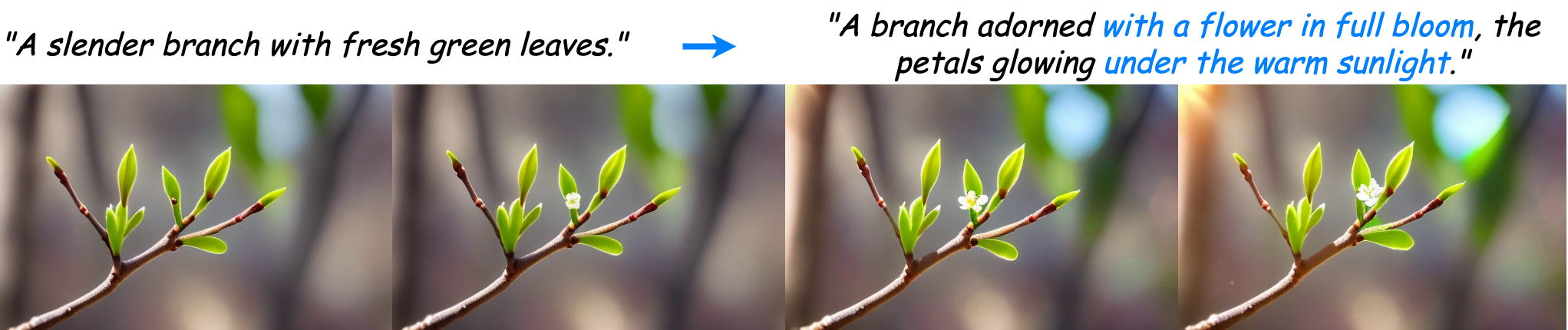}
}

\subfloat[Object Count Change.]{
\includegraphics[clip,width=\columnwidth]{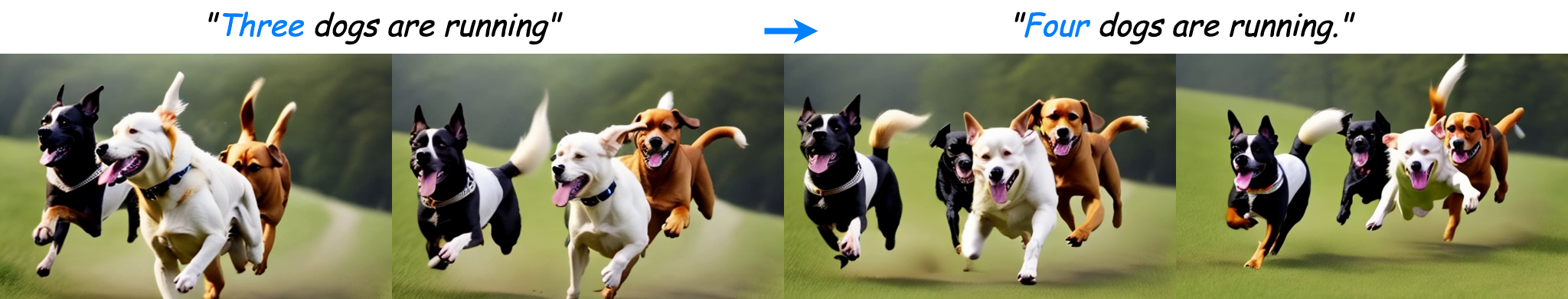}
}
\end{center}
\vspace{-6mm}
   \caption{More Applications. We showcase three enhanced capabilities of our method: (a) detailed attribute control, (b) multiple attribute control, and (c) object count change for dynamic scene modifications while maintaining consistency.} 
   
\label{fig:moreapp}
\vspace{-4mm}
\end{figure}

Beyond the attribute transitions systematically evaluated in our benchmark, we demonstrate a greater applicability of our method in Figure~\ref{fig:moreapp} by allowing more detailed prompts over attribute changes. First, our approach allows for precise control over transitions that go beyond basic changes, offering detailed modifications. Additionally, our method supports multiple attribute control, enabling the simultaneous adjustment of multiple attributes while ensuring smooth transitions. Finally, we present the change of object count, facilitating dynamic scene modifications while preserving motion consistency. These capabilities underscore the versatility and broader applicability of our method, making it suitable for more complex, real-world scenarios.

\subsection{Limitations}
Our method is compatible with any classifier-free guidance method, including both UNet-based and transformer-based diffusion models, enabling smooth transitions without sacrificing video quality. Experiments are reported in the Supplemental Material.
However, the effectiveness of the approach depends on the performance of the backbone model. If the backbone model struggles to represent certain attributes, creating precise transitions can become difficult. Nonetheless, the compatibility of our approach with any classifier-free guidance approach suggests that as backbone models continue to evolve, our method will benefit from these advancements, leading to even better results over time.

\section{Conclusion}
\vspace{-2mm}
We present a simple yet effective approach for video generation with smooth attribute transitions. Our method introduces a transitional direction to guide the sampling process during denoising, ensuring gradual and coherent transitions. Using neutral prompt anchoring, we mitigate unwanted changes and maintain consistency. Additionally, we propose CAT-Bench, a benchmark designed to evaluate complex scenes involving both attribute transitions and motion dynamics. To assess transition quality, we introduce the Wholistic Transition Score and the Frame-wise Transition Score, which measure correctness and smoothness. Experimental results demonstrate that our framework enables seamless attribute transitions, advancing video generation capabilities. In addition, our proposed metrics align closely with human evaluations, confirming their reliability in assessing transition quality.

\section*{Acknowledgements}
This work was supported in part by the Institute of Information \& Communications Technology Planning \& Evaluation (IITP) grant funded by the Korean Government (MSIT) (No. RS-2024-00457882, National AI Research Lab Project). It was also partially supported by the National Science and Technology Council (NSTC), Taiwan, under Grant NSTC-112-2628-E-002-033-MY4, and was financially supported by the Center of Data Intelligence: Technologies, Applications, and Systems, National Taiwan University (Grants 114L900901, 114L900902, and 114L900903), through the Featured Areas Research Center Program of the Higher Education Sprout Project, Ministry of Education, Taiwan.

{
    \small
    \bibliographystyle{ieeenat_fullname}
    \bibliography{main}

\begin{thebibliography}{28}
\providecommand{\natexlab}[1]{#1}
\providecommand{\url}[1]{\texttt{#1}}
\expandafter\ifx\csname urlstyle\endcsname\relax
  \providecommand{\doi}[1]{doi: #1}\else
  \providecommand{\doi}{doi: \begingroup \urlstyle{rm}\Url}\fi

\bibitem[Achiam et~al.(2023)Achiam, Adler, Agarwal, Ahmad, Akkaya, Aleman, Almeida, Altenschmidt, Altman, Anadkat, et~al.]{achiam2023gpt}
Josh Achiam, Steven Adler, Sandhini Agarwal, Lama Ahmad, Ilge Akkaya, Florencia~Leoni Aleman, Diogo Almeida, Janko Altenschmidt, Sam Altman, Shyamal Anadkat, et~al.
\newblock Gpt-4 technical report.
\newblock \emph{arXiv preprint arXiv:2303.08774}, 2023.

\bibitem[Balaji et~al.(2022)Balaji, Nah, Huang, Vahdat, Song, Zhang, Kreis, Aittala, Aila, Laine, Catanzaro, Karras, and Liu]{balaji2022ediff}
Yogesh Balaji, Seungjun Nah, Xun Huang, Arash Vahdat, Jiaming Song, Qinsheng Zhang, Karsten Kreis, Miika Aittala, Timo Aila, Samuli Laine, Bryan Catanzaro, Tero Karras, and Ming-Yu Liu.
\newblock ediff-i: Text-to-image diffusion models with an ensemble of expert denoisers.
\newblock \emph{arXiv preprint arXiv:2211.01324}, 2022.

\bibitem[Brack et~al.(2023)Brack, Friedrich, Hintersdorf, Struppek, Schramowski, and Kersting]{brack2023sega}
Manuel Brack, Felix Friedrich, Dominik Hintersdorf, Lukas Struppek, Patrick Schramowski, and Kristian Kersting.
\newblock Sega: Instructing text-to-image models using semantic guidance.
\newblock In \emph{NeurIPS}, 2023.

\bibitem[Cao et~al.(2023)Cao, Wang, Qi, Shan, Qie, and Zheng]{cao2023masactrl}
Mingdeng Cao, Xintao Wang, Zhongang Qi, Ying Shan, Xiaohu Qie, and Yinqiang Zheng.
\newblock Masactrl: Tuning-free mutual self-attention control for consistent image synthesis and editing.
\newblock In \emph{CVPR}, 2023.

\bibitem[Chen et~al.(2024)Chen, Zhang, Cun, Xia, Wang, Weng, and Shan]{chen2024videocrafter2}
Haoxin Chen, Yong Zhang, Xiaodong Cun, Menghan Xia, Xintao Wang, Chao Weng, and Ying Shan.
\newblock Videocrafter2: Overcoming data limitations for high-quality video diffusion models.
\newblock In \emph{CVPR}, 2024.

\bibitem[Dhariwal and Nichol(2021)]{dhariwal2021diffusion}
Prafulla Dhariwal and Alexander Nichol.
\newblock Diffusion models beat gans on image synthesis.
\newblock In \emph{NeurIPS}, 2021.

\bibitem[Feng et~al.(2024)Feng, Li, Saxon, Fu, Chen, and Wang]{feng2024tc}
Weixi Feng, Jiachen Li, Michael Saxon, Tsu-jui Fu, Wenhu Chen, and William~Yang Wang.
\newblock Tc-bench: Benchmarking temporal compositionality in text-to-video and image-to-video generation.
\newblock \emph{arXiv preprint arXiv:2406.08656}, 2024.

\bibitem[Ge et~al.(2022)Ge, Hayes, Yang, Yin, Pang, Jacobs, Huang, and Parikh]{ge2022long}
Songwei Ge, Thomas Hayes, Harry Yang, Xi Yin, Guan Pang, David Jacobs, Jia-Bin Huang, and Devi Parikh.
\newblock Long video generation with time-agnostic vqgan and time-sensitive transformer.
\newblock In \emph{ECCV}, 2022.

\bibitem[Guo et~al.(2024)Guo, Yang, Rao, Liang, Wang, Qiao, Agrawala, Lin, and Dai]{guo2023animatediff}
Yuwei Guo, Ceyuan Yang, Anyi Rao, Zhengyang Liang, Yaohui Wang, Yu Qiao, Maneesh Agrawala, Dahua Lin, and Bo Dai.
\newblock Animatediff: Animate your personalized text-to-image diffusion models without specific tuning.
\newblock In \emph{ICLR}, 2024.

\bibitem[He et~al.(2022)He, Yang, Zhang, Shan, and Chen]{he2022lvdm}
Yingqing He, Tianyu Yang, Yong Zhang, Ying Shan, and Qifeng Chen.
\newblock Latent video diffusion models for high-fidelity long video generation.
\newblock \emph{arXiv preprint arXiv:2211.13221}, 2022.

\bibitem[Hertz et~al.(2023)Hertz, Mokady, Tenenbaum, Aberman, Pritch, and Cohen-Or]{hertz2022prompt}
Amir Hertz, Ron Mokady, Jay Tenenbaum, Kfir Aberman, Yael Pritch, and Daniel Cohen-Or.
\newblock Prompt-to-prompt image editing with cross attention control.
\newblock In \emph{ICLR}, 2023.

\bibitem[Ho and Salimans(2021)]{ho2021classifierfree}
Jonathan Ho and Tim Salimans.
\newblock Classifier-free diffusion guidance.
\newblock In \emph{NeurIPS}, 2021.

\bibitem[Hong et~al.(2023)Hong, Ding, Zheng, Liu, and Tang]{hong2022cogvideo}
Wenyi Hong, Ming Ding, Wendi Zheng, Xinghan Liu, and Jie Tang.
\newblock Cogvideo: Large-scale pretraining for text-to-video generation via transformers.
\newblock In \emph{ICLR}, 2023.

\bibitem[Huang et~al.(2023)Huang, Feng, Shi, Xu, Yu, and Yang]{huang2024freebloom}
Hanzhuo Huang, Yufan Feng, Cheng Shi, Lan Xu, Jingyi Yu, and Sibei Yang.
\newblock Free-bloom: Zero-shot text-to-video generator with llm director and ldm animator.
\newblock In \emph{NeurIPS}, 2023.

\bibitem[Huang et~al.(2024)Huang, He, Yu, Zhang, Si, Jiang, Zhang, Wu, Jin, Chanpaisit, et~al.]{huang2024vbench}
Ziqi Huang, Yinan He, Jiashuo Yu, Fan Zhang, Chenyang Si, Yuming Jiang, Yuanhan Zhang, Tianxing Wu, Qingyang Jin, Nattapol Chanpaisit, et~al.
\newblock Vbench: Comprehensive benchmark suite for video generative models.
\newblock In \emph{CVPR}, 2024.

\bibitem[Lee et~al.(2024)Lee, Kwon, and Kim]{lee2024grid}
Taegyeong Lee, Soyeong Kwon, and Taehwan Kim.
\newblock Grid diffusion models for text-to-video generation.
\newblock In \emph{CVPR}, 2024.

\bibitem[Liu et~al.(2022)Liu, Li, Du, Torralba, and Tenenbaum]{liu2022compositional}
Nan Liu, Shuang Li, Yilun Du, Antonio Torralba, and Joshua~B Tenenbaum.
\newblock Compositional visual generation with composable diffusion models.
\newblock In \emph{ECCV}, 2022.

\bibitem[Ma et~al.(2024)Ma, Wang, Jia, Chen, Liu, Li, Chen, and Qiao]{ma2024latte}
Xin Ma, Yaohui Wang, Gengyun Jia, Xinyuan Chen, Ziwei Liu, Yuan-Fang Li, Cunjian Chen, and Yu Qiao.
\newblock Latte: Latent diffusion transformer for video generation.
\newblock \emph{arXiv preprint arXiv:2401.03048}, 2024.

\bibitem[Oh et~al.(2024)Oh, Jeong, Kim, Byeon, Kim, Kim, and Kim]{oh2025mevg}
Gyeongrok Oh, Jaehwan Jeong, Sieun Kim, Wonmin Byeon, Jinkyu Kim, Sungwoong Kim, and Sangpil Kim.
\newblock Mevg: Multi-event video generation with text-to-video models.
\newblock In \emph{ECCV}, 2024.

\bibitem[Qiu et~al.(2024)Qiu, Xia, Zhang, He, Wang, Shan, and Liu]{qiu2023freenoise}
Haonan Qiu, Menghan Xia, Yong Zhang, Yingqing He, Xintao Wang, Ying Shan, and Ziwei Liu.
\newblock Freenoise: Tuning-free longer video diffusion via noise rescheduling.
\newblock In \emph{ICLR}, 2024.

\bibitem[Shen et~al.(2020)Shen, Gu, Tang, and Zhou]{shen2020interpreting}
Yujun Shen, Jinjin Gu, Xiaoou Tang, and Bolei Zhou.
\newblock Interpreting the latent space of gans for semantic face editing.
\newblock In \emph{CVPR}, 2020.

\bibitem[Song et~al.(2021)Song, Durkan, Murray, and Ermon]{song2021score}
Yang Song, Conor Durkan, Iain Murray, and Stefano Ermon.
\newblock Maximum likelihood training of score-based diffusion models.
\newblock In \emph{NeurIPS}, 2021.

\bibitem[Tian et~al.(2024)Tian, Yang, Yang, Gao, Deng, Chen, Wang, Yu, Tao, Wan, et~al.]{tian2024videotetris}
Ye Tian, Ling Yang, Haotian Yang, Yuan Gao, Yufan Deng, Jingmin Chen, Xintao Wang, Zhaochen Yu, Xin Tao, Pengfei Wan, et~al.
\newblock Videotetris: Towards compositional text-to-video generation.
\newblock In \emph{NeurIPS}, 2024.

\bibitem[Villegas et~al.(2022)Villegas, Babaeizadeh, Kindermans, Moraldo, Zhang, Saffar, Castro, Kunze, and Erhan]{villegas2022phenaki}
Ruben Villegas, Mohammad Babaeizadeh, Pieter-Jan Kindermans, Hernan Moraldo, Han Zhang, Mohammad~Taghi Saffar, Santiago Castro, Julius Kunze, and Dumitru Erhan.
\newblock Phenaki: Variable length video generation from open domain textual descriptions.
\newblock In \emph{ICLR}, 2022.

\bibitem[Wang et~al.(2023{\natexlab{a}})Wang, Chen, Song, Ye, Liu, and Li]{wang2023genl}
Fu-Yun Wang, Wenshuo Chen, Guanglu Song, Han-Jia Ye, Yu Liu, and Hongsheng Li.
\newblock Gen-l-video: Multi-text to long video generation via temporal co-denoising.
\newblock \emph{arXiv preprint arXiv:2305.18264}, 2023{\natexlab{a}}.

\bibitem[Wang et~al.(2023{\natexlab{b}})Wang, Yuan, Chen, Zhang, Wang, and Zhang]{wang2023modelscope}
Jiuniu Wang, Hangjie Yuan, Dayou Chen, Yingya Zhang, Xiang Wang, and Shiwei Zhang.
\newblock Modelscope text-to-video technical report.
\newblock \emph{arXiv preprint arXiv:2308.06571}, 2023{\natexlab{b}}.

\bibitem[Wu et~al.(2023)Wu, Ge, Wang, Lei, Gu, Shi, Hsu, Shan, Qie, and Shou]{wu2023tune}
Jay~Zhangjie Wu, Yixiao Ge, Xintao Wang, Stan~Weixian Lei, Yuchao Gu, Yufei Shi, Wynne Hsu, Ying Shan, Xiaohu Qie, and Mike~Zheng Shou.
\newblock Tune-a-video: One-shot tuning of image diffusion models for text-to-video generation.
\newblock In \emph{ICCV}, 2023.

\bibitem[Zheng et~al.(2024)Zheng, Peng, Yang, Shen, Li, Liu, Zhou, Li, and You]{zheng2024open}
Zangwei Zheng, Xiangyu Peng, Tianji Yang, Chenhui Shen, Shenggui Li, Hongxin Liu, Yukun Zhou, Tianyi Li, and Yang You.
\newblock Open-sora: Democratizing efficient video production for all.
\newblock \emph{arXiv preprint arXiv:2412.20404}, 2024.

\end{thebibliography}
}


\clearpage
\setcounter{page}{1}
\maketitlesupplementary

\section{Implementation Details}
During sampling, the classifier-free guidance scale $\omega$ is set to 12, and the scaling factor $\alpha$ is configured to [0,1], ensuring the attribute transition scale ranges from -1 to 1 across frames in the video.

For quantitative evaluation on CAT-Bench, we generate five videos per prompt, yielding a total of 600 videos. Scores are averaged across all generated videos, and the default resolution settings of each model are used to ensure consistency. Similarly, for TC-Bench-T2V, we generate five videos per prompt and report the TCR and TC-Score using GPT-4~\cite{achiam2023gpt} for assertion-based evaluations. For competing methods, we reference the scores reported in the original paper~\cite{feng2024tc}. All models can be run on a single 24 GB NVIDIA RTX A5000.

For the user study, we combined our generated videos with those produced by three other baseline methods using the same backbone, ensuring the order was randomized to eliminate bias. A total of 38 participants were asked to evaluate the videos and select the best one based on five criteria: Transition Correctness, Transition Smoothness, Video Consistency, Motion Accuracy, and Overall Performance.

\section{Details of CAT-Bench}

\begin{table*}[!h]
\centering
\small
\caption{Examples of the Prompts in Our CAT-Bench. The \underline{underline} highlights the transitioning attribute in the prompt pair for multi-prompt generation.}
\label{tab:cat_ex}
\begin{tabular}{l|l}
\toprule
\myrowcoloursub
Multi-Prompt& Single-Prompt\\
\makecell[l]{\textit{``A \underline{young} man is rowing a boat''} \\$\rightarrow$
\textit{``An \underline{elderly} man is rowing a boat''}}&
\textit{``A man is rowing a boat, gradually aging from young to old over time.''}\\

\myrowcolour
\makecell[l]{\textit{``A woman \underline{without makeup} is laughing at the party. ''} \\$\rightarrow$
\textit{"A woman \underline{with makeup} is laughing at the party. "}}&
\makecell[l]{\textit{``A woman is laughing at the party, gradually transforming from having no makeup} \\
\textit{ to wearing makeup.''}}\\

\makecell[l]{\textit{"A man \underline{without a beard} is applauding. "} \\$\rightarrow$
\textit{``A man \underline{with a beard} is applauding. ''}}&
\textit{``A man is applauding, gradually transitioning from being clean-shaven to having a beard.''}\\

\myrowcolour
\makecell[l]{\textit{``A woman \underline{with short hair} is riding a horse.''} \\$\rightarrow$
\textit{``A woman \underline{with long hair} is riding a horse.''}}&
\textit{``A woman is riding a horse, with her hair gradually transitioning from short to long.''}\\

\makecell[l]{\textit{``A \underline{white} dog is running in the field.''} \\$\rightarrow$
\textit{``A \underline{gray} dog is running in the field.''}}&
\textit{``A dog is running in the field, gradually transitioning from white to gray.''}\\

\myrowcolour
\makecell[l]{\textit{``A \underline{knit} shirt is floating in the wind.''} \\$\rightarrow$
\textit{``A \underline{silk} shirt is floating in the wind.''}}&
\textit{``A shirt is floating in the wind, gradually transitioning from knit to silk.''}\\

\makecell[l]{\textit{``A boat is drifting on the river \underline{in a dark light.}''} \\$\rightarrow$
\textit{``A boat is drifting on the river \underline{in a bright light.}''}}&
\textit{``A boat is drifting on the river as the light transitions gradually from dark to bright.''}\\

\myrowcolour
\makecell[l]{\textit{``A hot air balloon is flying \underline{on a sunny day}.''} \\$\rightarrow$
\textit{``A hot air balloon is flying \underline{on a cloudy day}.''}}&
\textit{``A hot air balloon is flying across the sky as the weather transitions from sunny to cloudy.''}\\

\bottomrule
\end{tabular}
\end{table*}

We define 8 categories of attributes for transitions: four focused on human characteristics (age, beard, makeup, and hair), two on subject attributes (color and material), and two on background features (light conditions and weather). Each category includes 15 samples, resulting in a diverse dataset.
For a comprehensive evaluation, each sample includes a prompt pair for multi-prompt generation methods and a single prompt for single-prompt generation. We ensure that each sample incorporates both an attribute transition and consistent motion, allowing the evaluation to cover not only the successful completion of attribute transitions but also the preservation of motion dynamics. This ensures that the generated videos are not merely interpolations between static images without motion.

To construct the benchmark, we first define the prompt pairs and then use GPT-4 to generate a single descriptive prompt that encapsulates the attribute transition within a single prompt and facilitates fair and consistent evaluation across different methods. Table~\ref{tab:cat_ex} presents some examples from our CAT-Bench.

\section{Discussion on Video Quality}
\begin{table*}[ht]
\centering
\small
\caption{Quantitative Evaluation for Overall Video Quality. The highest scores are highlighted in \highest{blue} and the second-highest scores are highlighted in \secondhighest{magenta}. }
\label{tab:video_quality}
\begin{tabular}{lccccc}
\toprule
 & Consistency$\uparrow$ & Imaging Quality$\uparrow$ & Temporal Flickering$\uparrow$ & Motion Smoothness$\uparrow$ & Dynamic Degree$\uparrow$\\
\midrule

AnimateDiff& 0.8044 & 0.4848 & 0.7007 & 0.7309 & \highest{\textbf{1.0000}}\\
\myrowcolour
ModelScope& 0.9204 & 0.6075 & 0.9528 & 0.9651 & 0.5083\\

Latte& 0.9178 & 0.5600 & 0.9352 & 0.9523&\secondhighest{\underline{0.7808}}\\
\myrowcolour
Free-Bloom& 0.8422 & \highest{\textbf{0.7086}} & 0.8987 & 0.9091 & 0.4833\\

VideoTetris& 0.8905 & 0.5670 & 0.9337& 0.9535 & 0.6999\\
\midrule
\myrowcolour
VideoCrafter2& \highest{\textbf{0.9639}} & 0.6549 & \highest{\textbf{0.9595}} & 0.9774& 0.4750\\

Gen-L& 0.9527 & 0.6429 & 0.9525& \secondhighest{\underline{0.9776}} & 0.5749\\
\myrowcolour
FreeNoise& \secondhighest{\underline{0.9556}} & 0.6628 & 0.9547 & 0.9748 & 0.3888\\

\textbf{Ours}& 0.9537 & \secondhighest{\underline{0.6648}} & \secondhighest{\underline{0.9573}} & \highest{\textbf{0.9780}} & 0.7083\\
\bottomrule
\end{tabular}
\end{table*}
We employ metrics from VBench~\cite{huang2024vbench} to evaluate overall video quality across five key dimensions. First, we assess \textit{Consistency}. For human and object attribute transitions, we use background consistency to ensure that elements aside from the attribute remain stable across frames, while for background attribute transitions, we employ subject consistency. Next, we evaluate \textit{Temporal Flickering} to measure the overall temporal stability of the video. We then assess \textit{Motion Smoothness} to ensure that the generated movements are fluid and natural. As a complete static video could score well on previous metrics, we use \textit{Dynamic Degree}, which checks whether the video has sufficient dynamism.  Finally, we evaluate \textit{Imaging Quality} on a frame-by-frame basis to ensure high visual fidelity throughout the video.
Table~\ref{tab:video_quality} presents the evaluation results for video quality. The proposed approach achieves comparable scores to the backbone model, VideoCrafter2, across metrics such as Consistency, Imaging Quality, Temporal Flickering, and Motion Smoothness, indicating that our method maintains video quality while generating attribute transitions. Additionally, our approach achieves the highest score in Dynamic Degree among approaches using VideoCrafter2 as the backbone, demonstrating that the motion described in the prompts is effectively captured. Although AnimateDiff attains the highest Dynamic Degree, its low Motion Smoothness suggests that the high Dynamic score results from frame inconsistencies rather than genuine motion dynamics. In contrast, our method strikes a balance, ensuring both smooth motion with fluidity.

\section{Discussion on Scale Factor $\alpha$}
\begin{figure}
\begin{center}
\includegraphics[width=\columnwidth]{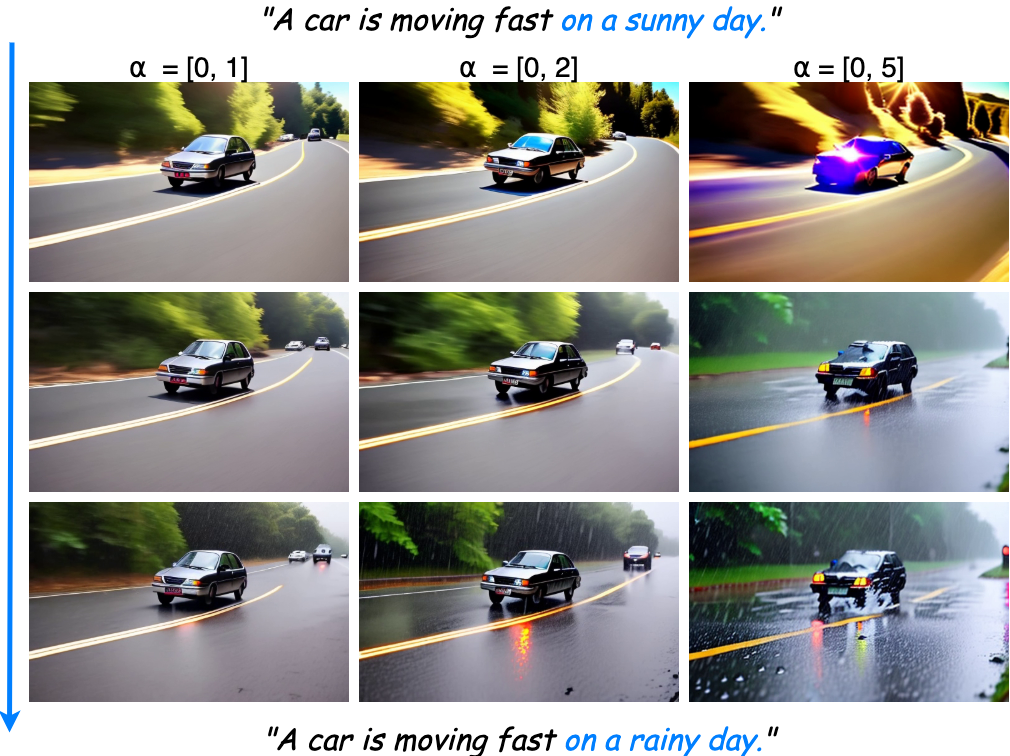}
\end{center}
   \caption{Effect of Scaling Factor ($\alpha$) on Attribute Transitions. Increasing $\alpha$ intensifies the transition (e.g., heavier rain, brighter sun), but excessively large values (e.g., $\alpha = 5$) introduce artifacts.} 
\label{fig: supp_scale}
\end{figure}
Unlike previous approaches that rely on prompt interpolation or leverage LLMs to define intermediate attribute states, our method not only facilitates attribute transitions but also enables control over the intensity of these transitions through the scaling factor, $\alpha$. As illustrated in Figure~\ref{fig: supp_scale}, setting $\alpha$ to [0,1], corresponding to a transition range from -1 to 1 across the video, results in a well-balanced progression from sunny to rainy. 
On the other hand, increasing $\alpha$ to [0,2], representing a range from -2 to 2, intensifies the transition. The rain in the final frame becomes heavier, while the sunlight in the initial frame shines more brightly, amplifying the visual impact of the attribute change. The flexibility highlights the adaptability of our approach in controlling transition magnitude to suit different scenarios.

While our method performs well with different $\alpha$ values, excessively large values can still lead to the distance effect. For instance, when $\alpha$ is set to [0,5], the video begins to exhibit significant artifacts and distortions, which occurs because driving the latent too far from the decision boundary leads to unnatural changes and inconsistencies, as mentioned earlier in the main paper.

\section{Discussion on Compatibility}
We demonstrate the compatibility of our method by applying it to OpenSora~\cite{zheng2024open}, a transformer-based model. As shown in Table~\ref{tab: compatibility}, we evaluate both OpenSora and our method, using OpenSora as the base model, on CAT-Bench. Our evaluation considers both attribute transition quality and overall video quality. The results indicate that our approach successfully enables OpenSora to generate smooth attribute transitions while maintaining high video quality. This experiment highlights the flexibility of our method, demonstrating its effectiveness even with transformer-based architectures.

\begin{table}[!h]
\caption{evaluation on compatibility}
\label{tab: compatibility}
\begin{tabular}{l|c|c}
\toprule
& OpenSora & Ours\\ 
\midrule
\myrowcolour
Motion Smoothness $\uparrow$ & 0.9723 & \highest{\textbf{0.9809}}\\
Dynamic Degree $\uparrow$& 0.4622 &\highest{\textbf{0.7194}}\\
\myrowcolour
Temporal Flickering$\uparrow$ & 0.9542 & \highest{\textbf{0.9566}}\\
Imaging Quality $\uparrow$ &0.6632&\highest{\textbf{0.6751}}\\
\myrowcolour
Background Consistency $\uparrow$& \highest{\textbf{0.9701 }}& 0.9602\\
\midrule
Wholistic Transition Score $\uparrow$ & 0.0035 &\highest{\textbf{0.1622}}\\
\myrowcolour
Frame-wise Transition Score $\uparrow$ & 0.0002 & \highest{\textbf{0.0193}}\\
\bottomrule

\end{tabular}
\end{table}

\section{Discussion on Inference Time}
\begin{figure}
\begin{center}
\includegraphics[width=\columnwidth]{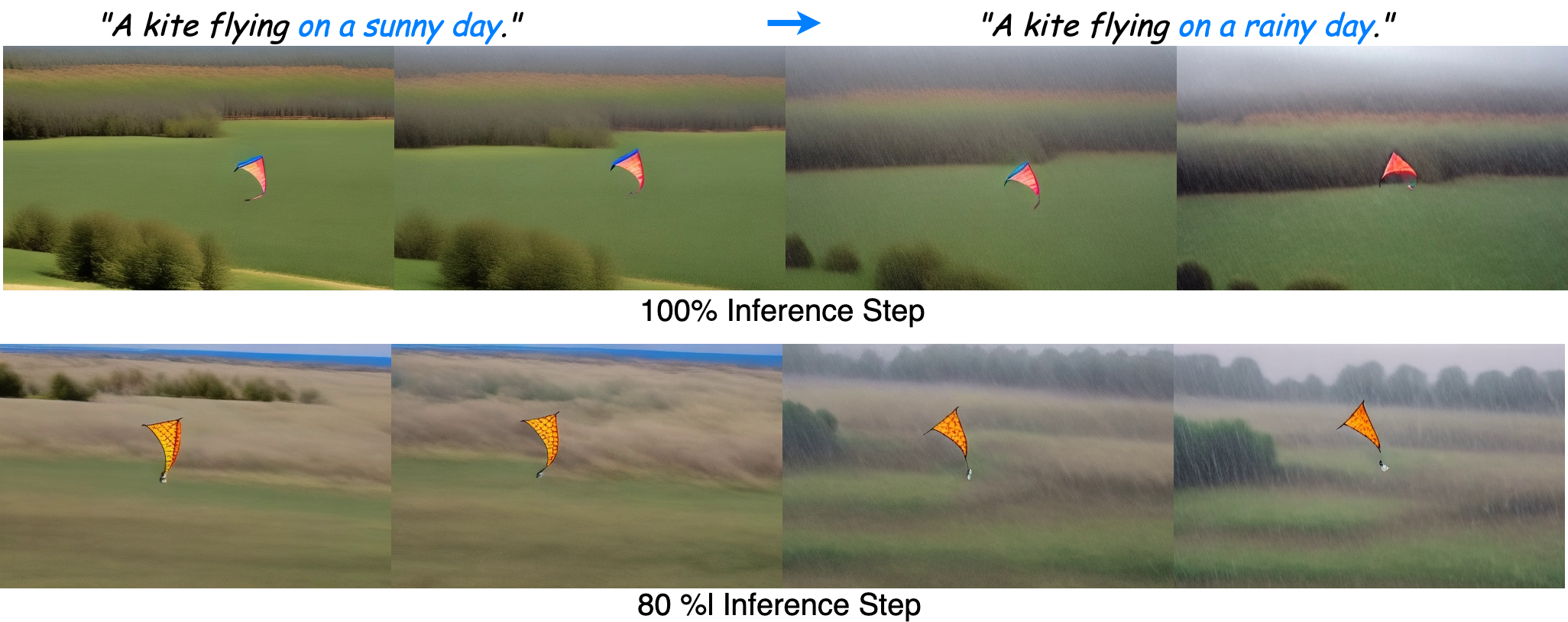}
\end{center}
\vspace{-4mm}
   \caption{Video quality comparison with reduced sampling steps. Despite using only 80\% of the sampling steps, the generated videos maintain similar visual quality and smooth attribute transitions.} 
\label{fig:inference_time}
\vspace{-4mm}
\end{figure}

We evaluate the inference time of our method and compare it to the base model, as shown in Table~\ref{tab:inference_time}. Our method requires inference time slightly higher than the base model, which is a reasonable trade-off for enabling smooth attribute transitions. Additionally, by reducing the sampling steps to 80\%, we achieve an inference time of 349.23 seconds, which is nearly on par with the base model while still maintaining smooth transitions. Furthermore, Figure~\ref{fig:inference_time} illustrates that reducing the sampling steps does not significantly degrade video quality, demonstrating that our approach remains effective even with fewer denoising iterations.

\begin{table}[h]
\small
\centering
\caption{Inference time comparison between the base model and our method.}
\begin{tabular}{l c c}
    \toprule
    Method & Sampling Steps & Inference Time (sec) \\
    \midrule
    Base Model & 100\% & 336.36 \\
    Ours (Full Steps) & 100\% & 453.71 \\
    Ours (Reduced Steps) & 80\% & 349.23 \\
    \bottomrule
\end{tabular}
\label{tab:inference_time}
\end{table}

\section{Additional Results}
\begin{figure*}
\begin{center}
\includegraphics[scale =0.3]{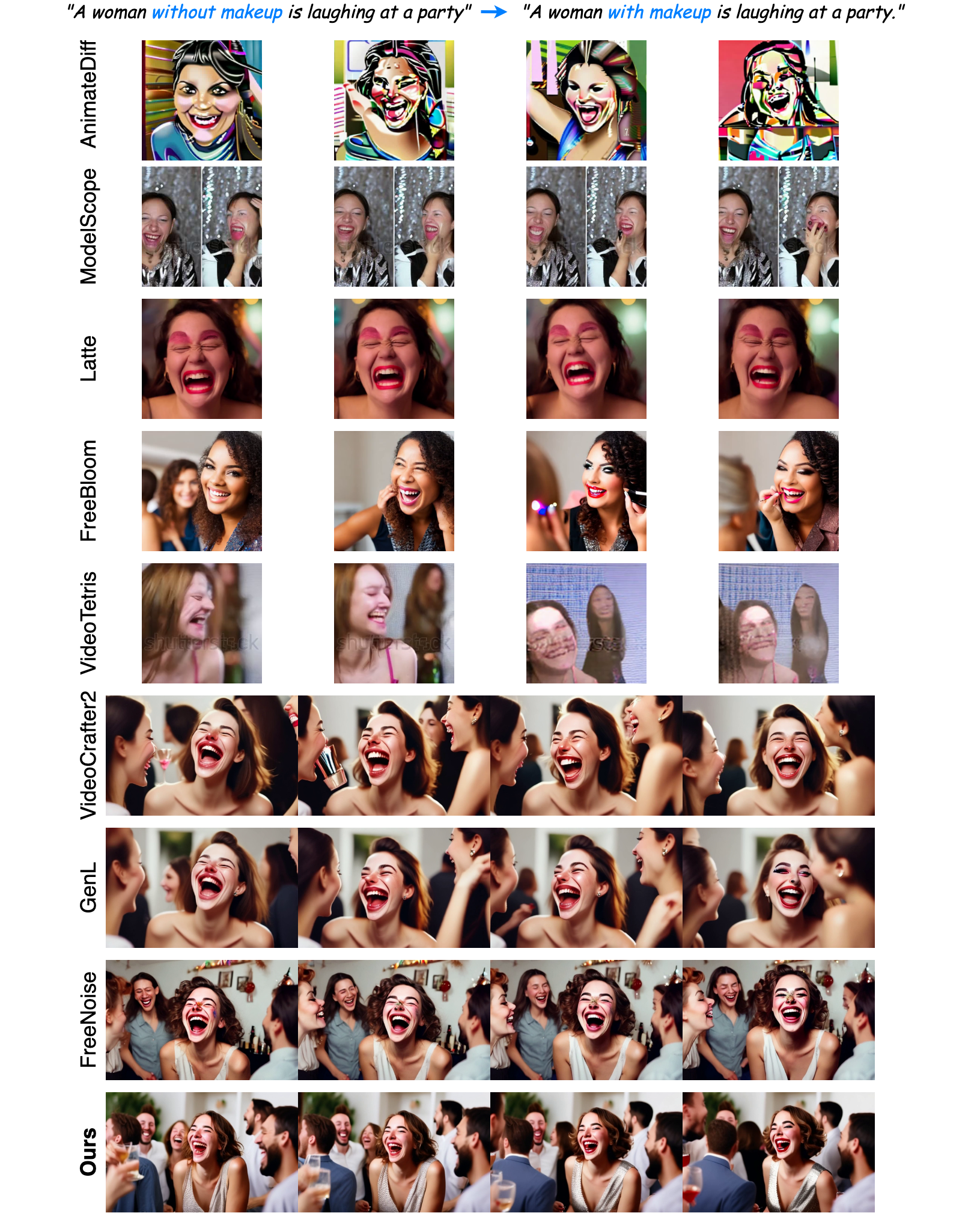}
\end{center}
   \caption{More Generation Results for the Specified Attribute Transition.} 
\label{fig: supp_more_1}
\end{figure*}

\begin{figure*}
\begin{center}
\includegraphics[scale =0.3]{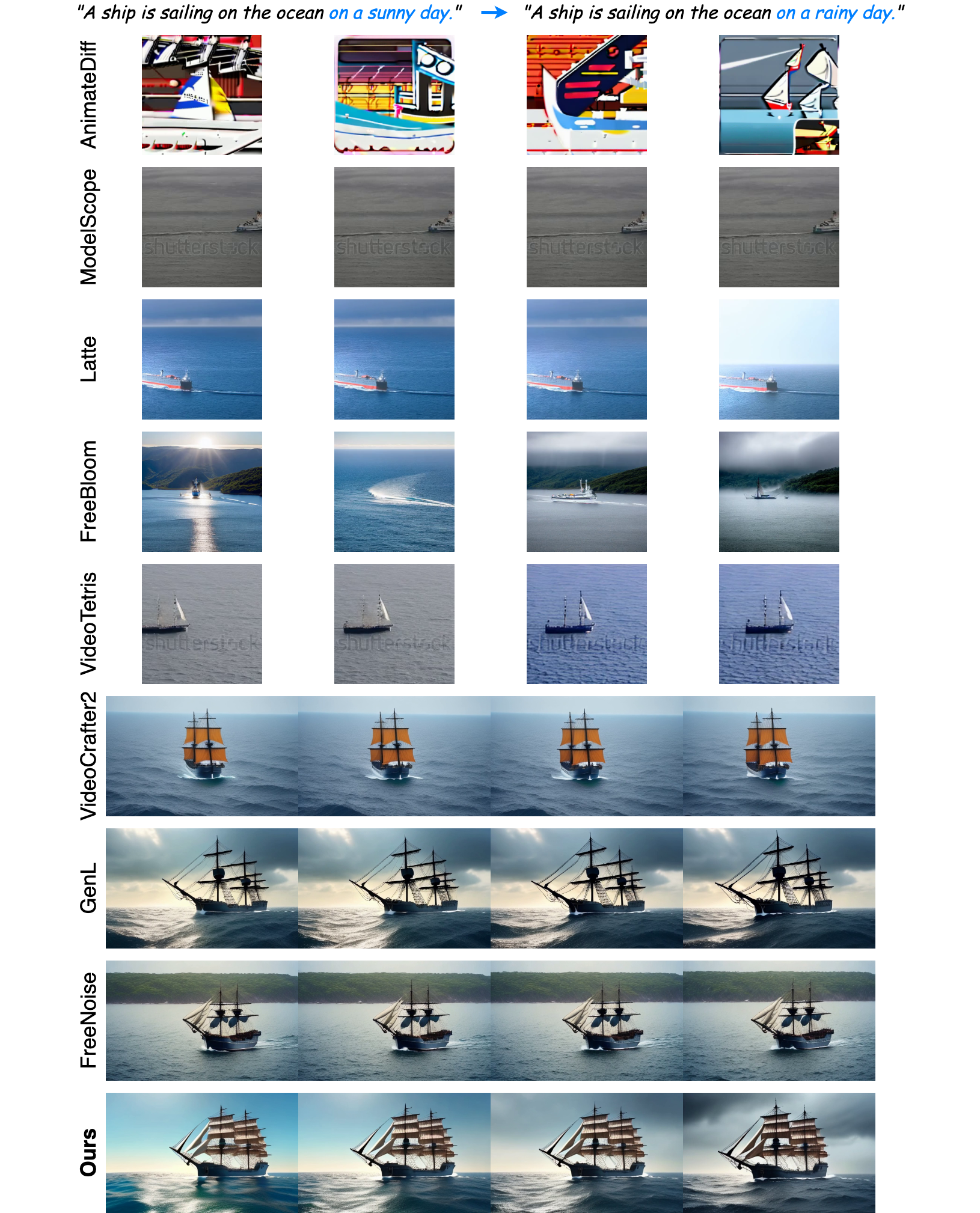}
\end{center}
   \caption{More Generation Results for the Specified Attribute Transition.} 
\label{fig: supp_more_2}
\end{figure*}

\begin{figure*}
\begin{center}
\includegraphics[scale =0.3]{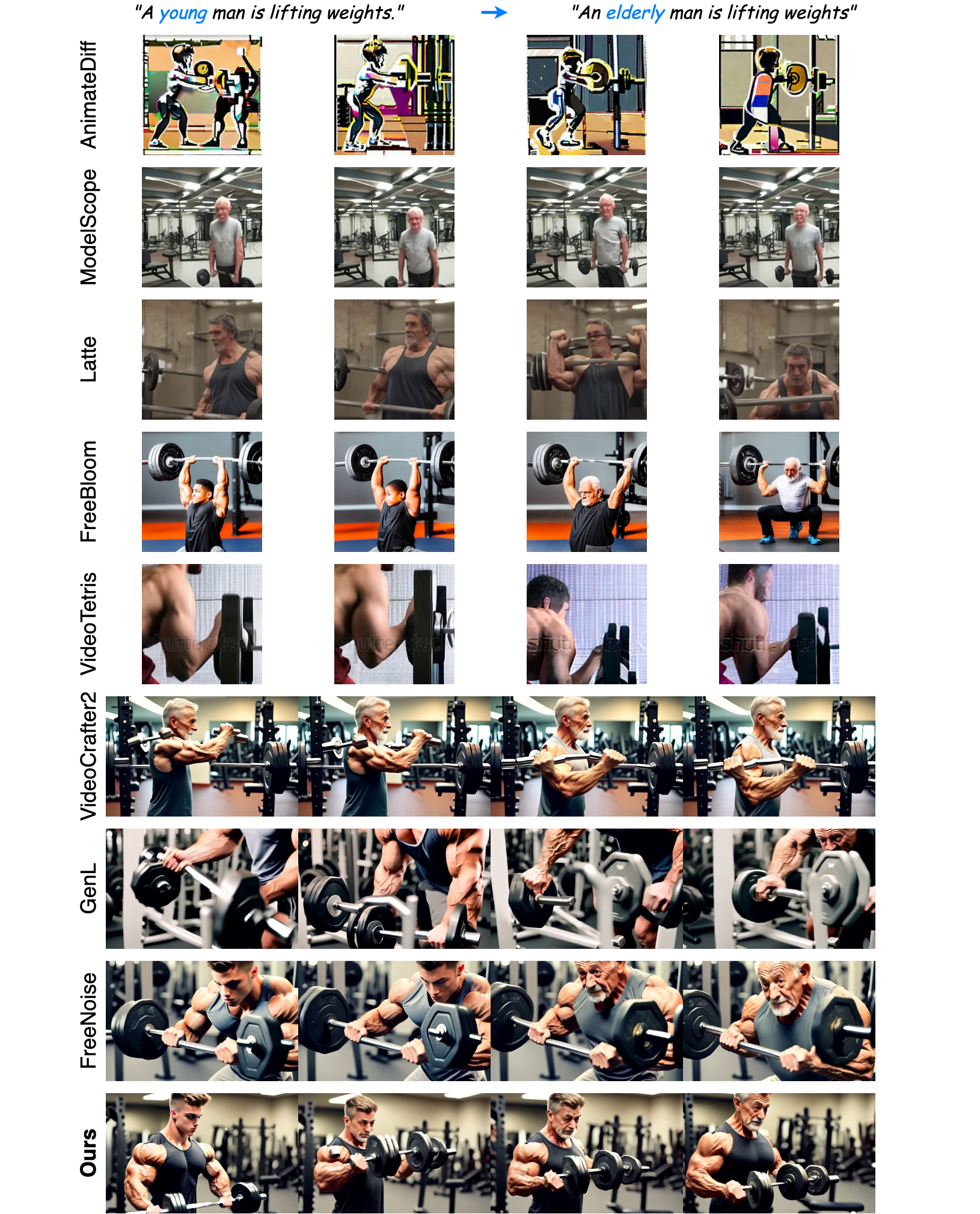}
\end{center}
   \caption{More Generation Results for the Specified Attribute Transition.} 
\label{fig: supp_more_3}
\end{figure*}

\begin{figure*}
\begin{center}
\includegraphics[scale =0.3]{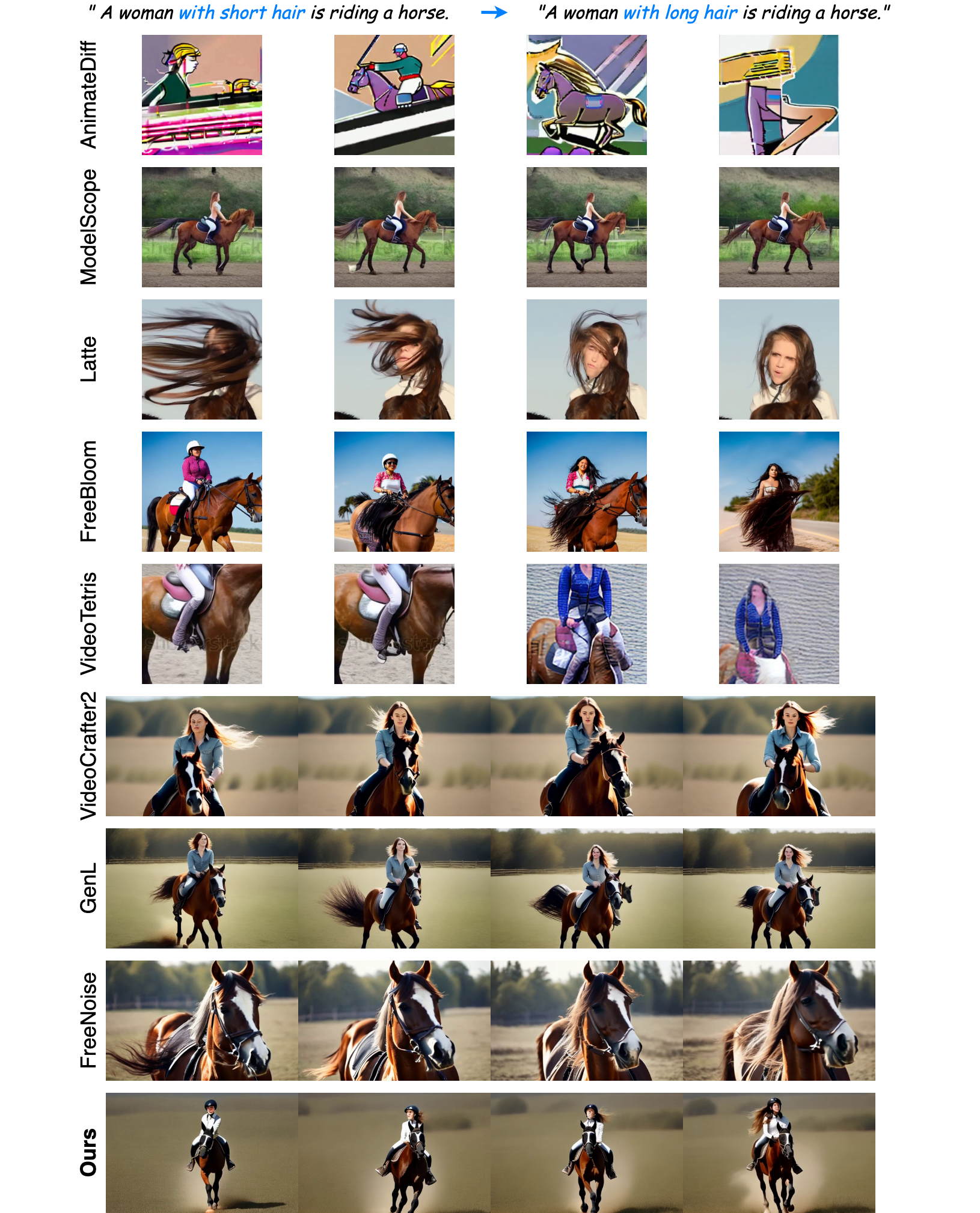}
\end{center}
   \caption{More Generation Results for the Specified Attribute Transition.} 
\label{fig: supp_more_4}
\end{figure*}

We present additional qualitative evaluation examples in Figures~\ref{fig: supp_more_1}-\ref{fig: supp_more_4} to further demonstrate the effectiveness of our method in generating smooth and accurate attribute transitions compared to baseline approaches. Additionally, we include video examples as part of the supplementary material for a more comprehensive evaluation of the performance of our approach.

\section{Longer Video with Attribute Transition}
\begin{figure*}
\begin{center}
\includegraphics[scale =0.2]{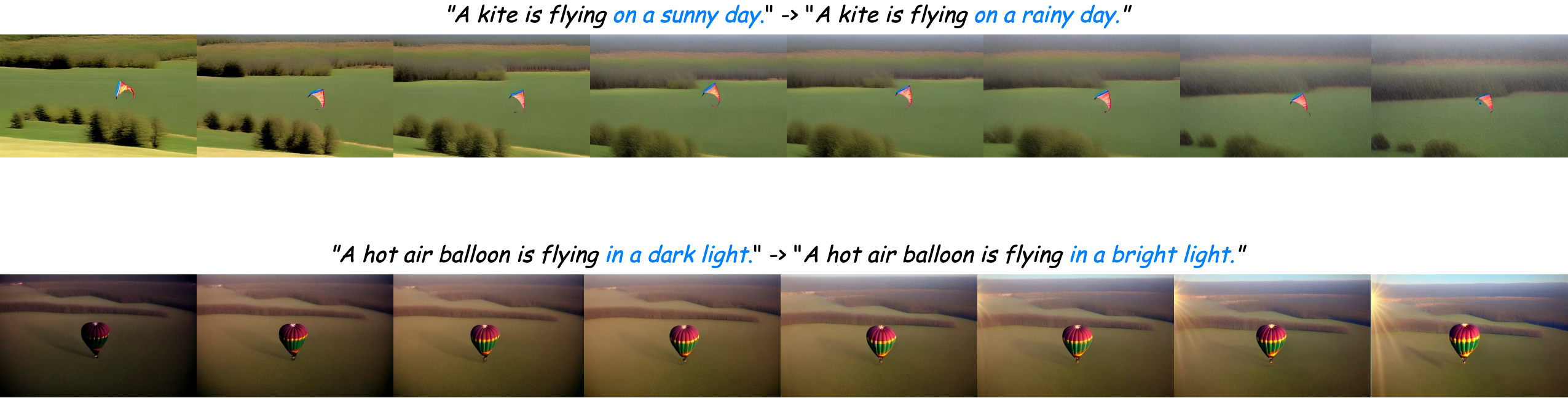}
\end{center}
   \caption{Generation Results of Longer Video.} 
\label{fig: supp_longer}
\end{figure*}

The proposed method is compatible with any baseline model utilizing classifier-free guidance. Additionally, it can seamlessly integrate with other sampling techniques for generating extended videos, enabling the creation of longer and more dynamic video sequences. To demonstrate its versatility, we integrate Temporal Co-Denoising from~\cite{wang2023genl} with our method and generate 64 frames per video. The results in Figure~\ref{fig: supp_longer} highlight the capability of our method to combine effectively with longer video generation techniques, producing consistent videos with smooth and accurate attribute transitions.  We provide video examples alongside the supplementary material document.

\end{document}